\documentclass[10pt,twocolumn,letterpaper]{article}

\usepackage{iccv}
\usepackage{times}
\usepackage{epsfig}
\usepackage{graphicx}
\usepackage{amsmath}
\usepackage{amssymb}
\usepackage{tikz}
\usepackage{caption}
\usepackage{subcaption}
\usepackage{float}
\usepackage{tikz} %pagenodes}
\usetikzlibrary{calc,positioning} % <----

\usepackage{bm,upgreek}      % for \bm macro
\usepackage{caption}

\newcommand{\RealBenchmarkName}{\emph{EgoDexter}}

\newcommand{\parahead}[1]{\noindent\textbf{#1}:\ }

\newenvironment{packed_itemize}
{\begin{itemize}
    \setlength{\itemsep}{1pt}
    \setlength{\parskip}{0pt}
    \setlength{\parsep}{0pt}
}{\end{itemize}}

\newcommand{\croppedinput}{\widetilde{\mathcal{D}}}

\newcommand{\filluptopage}[1]{%
  \clearpage
  \loop\ifnum\value{page}<#1\relax
    \null\clearpage
  \repeat
  \loop\ifnum\value{page}=#1\relax
    \null\clearpage
  \repeat
}

\newcommand{\cref}[1]{Section \ref{#1}}

\usepackage[pagebackref=true,breaklinks=true,colorlinks,bookmarks=false]{hyperref}

\iccvfinalcopy % *** Uncomment this line for the final submission

\ificcvfinal\fi

\begin{document}

%%%%%%%%% TITLE
\title{Real-time Hand Tracking under Occlusion from an Egocentric \mbox{RGB-D} Sensor}

\author{
	\parbox{5cm}{\centering Franziska Mueller$^{1}$\\Srinath Sridhar$^{1}$}\quad
	\parbox{5cm}{\centering Dushyant Mehta$^{1}$\\Dan Casas$^{2}$}\quad
	\parbox{5cm}{\centering Oleksandr Sotnychenko$^{1}$\\Christian Theobalt$^{1}$}\\[1.2em] 
	$^{1}$ Max-Planck-Institute for Informatics, Germany\quad
	$^{2}$ Universidad Rey Juan Carlos, Spain\quad
}

\maketitle

\begin{abstract}
We present an approach for real-time, robust and accurate hand pose estimation from moving egocentric \mbox{RGB-D} cameras in cluttered real environments.
Existing methods typically fail for hand-object interactions in cluttered scenes imaged from egocentric viewpoints---common for virtual or augmented reality applications.
Our approach uses two subsequently applied Convolutional Neural Networks (CNNs) to localize the hand and regress 3D joint locations.
Hand localization is achieved by using a CNN to estimate the 2D position of the hand center in the input, even in the presence of clutter and occlusions.
The localized hand position, together with the corresponding input depth value, is used to generate a normalized cropped image that is fed into a second CNN to regress relative 3D hand joint locations in real time.
For added accuracy, robustness and temporal stability, we refine the pose estimates using a kinematic pose tracking energy.
To train the CNNs, we introduce a new photorealistic dataset that uses a merged reality approach to capture and synthesize large amounts of annotated data of natural hand interaction in cluttered scenes.
Through quantitative and qualitative evaluation, we show that our method is robust to self-occlusion and occlusions by objects, particularly in moving egocentric perspectives.
%\franzi{TODO: add numbers from evaluation when we have them}
\end{abstract}

%% main BODY TEXT
\section{Introduction}
\label{sec:intro}
Estimating the full articulated 3D pose of hands is becoming increasingly important due to the central role that hands play in everyday human activities.
Applications in activity recognition~\cite{rohrbach_cvpr2012}, motion control~\cite{zhao_tog2013}, human--computer interaction~\cite{sridhar_chi2015}, and virtual/augmented reality (VR/AR) require real-time and accurate hand pose estimation under challenging conditions.
Spurred by recent developments in commodity depth sensing, several methods that use a single RGB-D camera have been proposed~\cite{taylor_siggraph2016,sridhar_cvpr2015,tagliasacchi_sgp2015,qian_cvpr2014,ge_cvpr2016,wan2017crossing}.
In particular, methods that use Convolutional Neural Networks (CNNs), possibly in combination with model-based hand tracking, have been shown to work well for static, third-person viewpoints in uncluttered scenes~\cite{tompson_tog2014,sinha_cvpr2016,oberweger_iccv2015}, \ie, mostly for free hand motion in mid-air, a setting that is uncommon in natural hand interaction.

\begin{figure}[t]
\centering
\includegraphics[width=1\columnwidth]{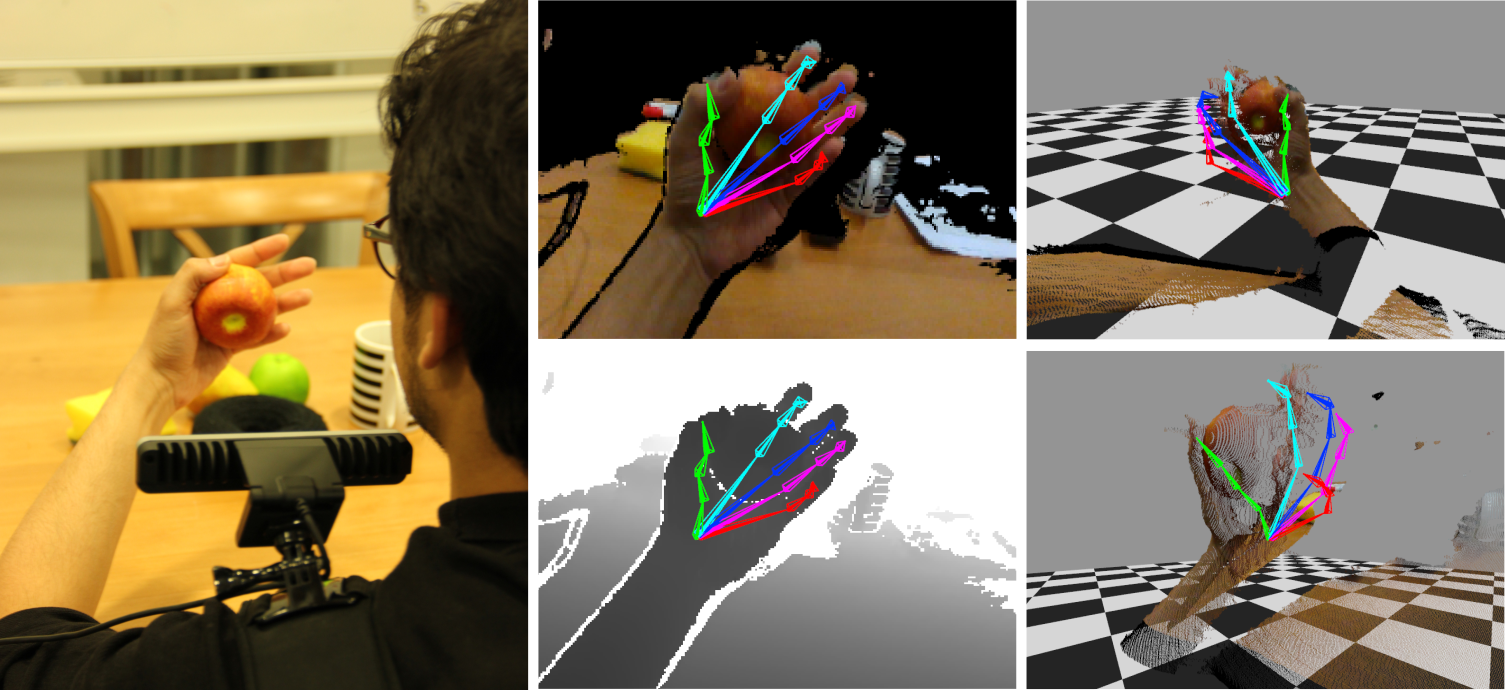}
\captionof{figure}{We present an approach to track the full 3D pose of the hand from egocentric viewpoints (left), a challenging problem due to additional self-occlusions, occlusions from objects and background clutter. Our method can reliably track the hand in 3D even under such conditions using only \mbox{RGB-D} input. Here we show tracking results overlaid with color and depth map (center), and visualized from virtual viewpoints (right). 
}
\label{fig:teaser}
\end{figure}

However, real-time hand pose estimation from \textbf{moving}, \textbf{first-person} camera viewpoints in \textbf{cluttered real-world scenes} where the hand is often occluded as it naturally interacts with objects, remains an unsolved problem.
We define first-person or \textbf{egocentric} viewpoints as those that would typically be imaged by cameras mounted on the head (for VR/AR applications), shoulder, or chest (see Figure~\ref{fig:teaser}).
Occlusions, cluttered backgrounds, manipulated objects, and field-of-view limitations make this scenario particularly challenging.
CNNs are a promising method to tackle this problem but typically require large amounts of \emph{annotated} data which is hard to obtain since markerless hand tracking (even with multiple views), and manual annotation on a large scale is infeasible in egocentric settings due to \mbox{(self-)}occlusions, cost, and time.
Even semi-automatic annotation approaches~\cite{oberweger_cvpr2016} would fail if large parts of the hand are occluded.
Photorealistic synthetic data, on the other hand, is inexpensive, easier to obtain, removes the need for manual annotation, and can produce accurate ground truth even under occlusion.

In this paper, we present, to our knowledge, the first method that supports \textbf{real-time} egocentric hand pose estimation in real scenes with cluttered backgrounds, occlusions, and complex hand-object interactions using a single commodity \mbox{RGB-D} camera.
We divide the task of per-frame hand pose estimation into: (1) hand localization, and (2) 3D joint location regression.
Hand localization, an important task in the presence of scene clutter, is achieved by a CNN that estimates the 2D image location of the center of the hand.
Further processing results in an image-level bounding box around the hand and the 3D location of the hand center (or of the occluding object directly in front of the center).
This output is fed into a second CNN that regresses the relative 3D locations of the 21 hand joints.
Both CNNs are trained with large amounts of fully annotated data which we obtain by combining hand-object interactions with real cluttered backgrounds using a new \textbf{merged reality} approach.
This increases the realism of the training data since users can perform motions to mimic manipulating a virtual object using live feedback of their hand pose.
These motions are rendered from novel egocentric views using a framework that photorealistically models \mbox{RGB-D} data of hands in natural interaction with objects and clutter.

The 3D joint location predictions obtained from the CNN are accurate but suffer from kinematic inconsistencies and temporal jitter expected in single frame pose estimation methods.
To overcome this, we refine the estimated 3D joint locations using a fast inverse kinematics pose tracking energy that uses 3D and 2D joint location constraints to estimate the joint angles of a temporally smooth skeleton.
Together, this results in the first real-time approach for smooth and accurate hand tracking even in cluttered scenes and from moving egocentric viewpoints.
We show the accuracy, robustness, and generality of our approach on a new benchmark dataset with moving egocentric cameras in real cluttered environments.
In sum, our contributions are:
\begin{packed_itemize}
\item A novel method that localizes the hand and estimates, in real time, the 3D joint locations from egocentric viewpoints, in clutter, and under strong occlusions using two CNNs. A kinematic pose tracking energy further refines the pose by estimating joint angles of a temporally smooth tracking.
\item A photorealistic data generation framework for synthesizing large amounts of annotated \mbox{RGB-D} training data of hands in natural interaction with objects and clutter.
\item Extensive evaluation on our new annotated real benchmark dataset \RealBenchmarkName~featuring egocentric cluttered scenes and interaction with objects.
\end{packed_itemize}

\section{Related work}
\label{sec:related-work}
Hand pose estimation has a rich history due to its applications in human--computer interaction, motion control and activity recognition.
However, most previous work estimates hand pose in mid-air and in uncluttered scenes with third-person viewpoints, making occlusions less of an issue.
We first review the prior art for this simpler setting (\emph{free hand tracking}) followed by a discussion of work in the harder hand-object and egocentric settings.

\parahead{Free Hand Tracking}
Many approaches for free hand tracking resort to multiple RGB cameras to overcome self-occlusions and achieve high accuracy~\cite{sridhar_iccv2013,ballan_eccv2012,wang20116d}.
However, single depth or \mbox{RGB-D} cameras are preferred since multiple cameras are cumbersome to setup and use.
Methods that use generative pose tracking have been successful for free hand tracking with only an \mbox{RGB-D} stream~\cite{oikonomidis2011efficient,qian_cvpr2014,tagliasacchi_sgp2015,tang_iccv2015}.
However, these approaches fail under typical fast motions, and occlusions due to objects and clutter.
To overcome this, most recent approaches rely solely on learning-based methods or combine them with generative pose tracking.
Random forests are a popular choice~\cite{kesin_iccvw2011,tang2014latent,sun2015cascaded,xu2013efficient,wan_eccv2016} due to their capacity but still result in kinematically inconsistent and jittery pose estimates.
Many methods overcome this limitation through combination with a generative pose tracking strategy~\cite{sridhar_cvpr2015,qian_cvpr2014,taylor_siggraph2016}.
All of the above approaches fail to work under occlusions arising from objects and scene clutter.
Recent deep learning methods promise large learning capacities for hand pose estimation~\cite{tompson_tog2014, ge_cvpr2016, sinha_cvpr2016,zhou_ijcai2016, ye_eccv2016,oberweger_iccv2015}.
However, generating enough examples for supervised training remains a challenge.
Commercial systems that claim to work for egocentric viewpoints \cite{leap_motion}
fail under large occlusions, see \cref{sec:results-and-evaluation}.

\parahead{Hand Tracking under Challenging Conditions}
Hand pose estimation under challenging scene, background, and camera conditions different from third-person mid-air tracking remains an unsolved problem.
Some methods can track hands even when they interact with objects~\cite{hamer2009tracking,romero_icra2010}, but they are limited to slow motions and limited articulation.
A method for real-time joint tracking of hands and objects from third-person viewpoints was recently proposed~\cite{sridhar_eccv2016}, but is limited to known objects and small occlusions. % \Dan{trained on non-natural?}
Methods for capturing complex hand-object interactions and object scanning were proposed~\cite{oikonomidis2011full,ballan_eccv2012,kyriazis_cvpr2014,tzionas20153d,tzionas_ijcv2016,panteleris20153d}.
However, these are offline methods and their performance in egocentric cluttered scenarios is unknown.
\begin{figure*}[!ht]
	\centering
	\includegraphics[width=1\textwidth]{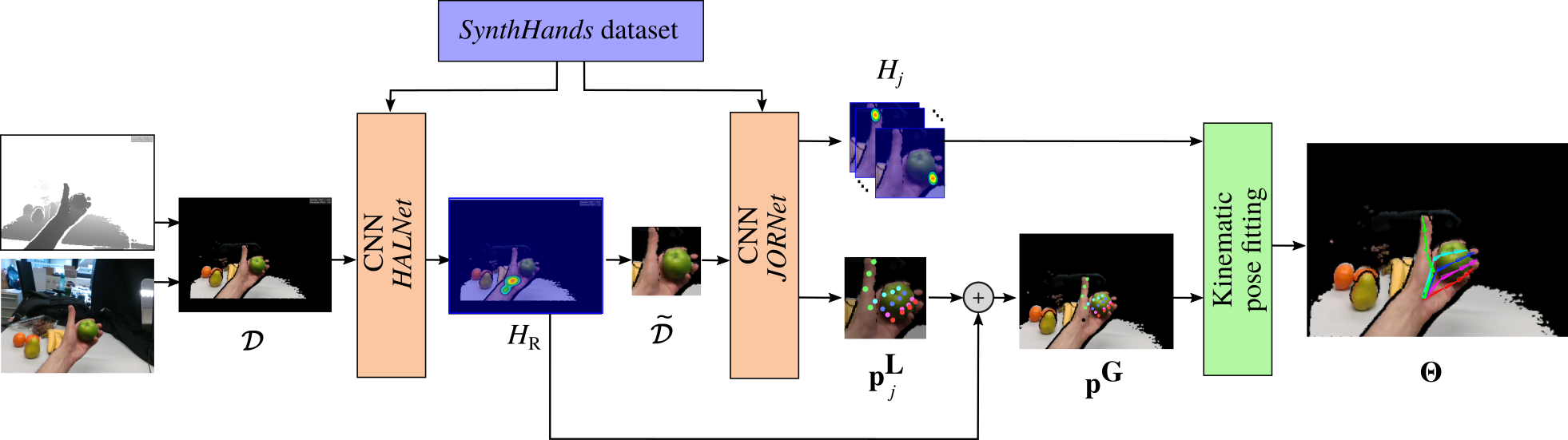}
	\caption{Overview: Starting from an \mbox{RGB-D} frame, we initially regress the 2D hand position heatmap using our CNN \emph{HALNet} and compute a cropped frame.
    A second CNN, \textit{JORNet}, is used to predict root-relative 3D joint positions as well as 2D joint heatmaps.
    Both CNNs are trainned with our new \emph{SynthHands} dataset.
    Finally, we use a pose tracking step to obtain the joint angles of a kinematic skeleton.}
	\label{fig:pipeline}
\end{figure*}

Using egocentric cameras for human performance capture has gained attention due to ready availability of consumer wearable cameras~\cite{rhodin2016egocap}.
Sridhar \etal~\cite{sridhar_cvpr2015} showed a working example of real-time egocentric tracking in uncluttered scenes.
Rogez \etal~\cite{rogez_eccv2014workshop,rogez_cvpr2015} presented one of the first methods to achieve this in cluttered scenes with natural hand-object interactions pioneering the use of synthetic images for training a machine learning approach for difficult egocentric views.
However, this work was not meant for real-time tracking.
We introduce an approach to leverage large amounts of synthetic training data
to achieve \emph{real-time}, temporally consistent hand tracking, even under challenging occlusion conditions.

\section{Overview}
\label{sec:overview}
Our goal is to estimate the full 3D articulated pose of the hand imaged with a single commodity \mbox{RGB-D} sensor.
We use the RGB and depth channels from the Intel RealSense SR300~\cite{intel_realsense_sr300}, both with a resolution of 640$\times$480 pixels and captured at 30~Hz.
We formulate hand pose estimation as an energy minimization problem that incorporates per-frame pose estimates into a temporal tracking framework.
The goal is to find the joint angles of a kinematic hand skeleton (\cref{sec:hand-model}) that best represent the input observation.
Similar strategies have been shown to be successful in state-of-the-art methods~\cite{taylor_siggraph2016,sridhar_cvpr2015,sridhar_eccv2016,qian_cvpr2014} that use per-frame pose estimation to initialize a tracker that refines and regularizes the joint angles of a kinematic skeleton for free hand tracking.
However, the per-frame pose estimation components of these methods struggle under strong occlusions, hand-object interactions, scene clutter, and moving egocentric cameras.
We overcome this limitation by combining a CNN-based 3D pose regression framework, that is tailored for this challenging setting, with a kinematic skeleton tracking strategy for temporally stable results.

We divide the task of hand pose estimation into several subtasks (Figure~\ref{fig:pipeline}).
First, \emph{hand localization} (\cref{sec:hand-localization}) is achieved by a CNN that estimates an image-level heatmap (that encodes position probabilities) of the \textbf{root} --- a point which is either the hand center (knuckle of the middle finger, shown with a star shape in Figure~\ref{fig:hand-model-3dpos}) or a point on an object that occludes the hand center.
The 2D and 3D root positions are used to extract a normalized cropped image of the hand.
Second, \emph{3D joint regression} (\cref{sec:3d-joint-regression}) achieved with a CNN that regresses root-relative 3D joint locations from the cropped hand image.
Both CNNs are trained with large amounts of annotated data which were generated with a novel framework to automatically generate 3D hand joint motion with natural hand interaction (\cref{sec:training}).
Finally, the regressed 3D joint positions are used in a kinematic pose tracking framework (\cref{sec:hand-pose-opt}) to obtain temporally smooth tracking of the hand motion.

\subsection{Hand Model}
\label{sec:hand-model}
\begin{figure}
	\centering
    \begin{subfigure}[t]{0.45\columnwidth}
        \includegraphics[width=\textwidth]{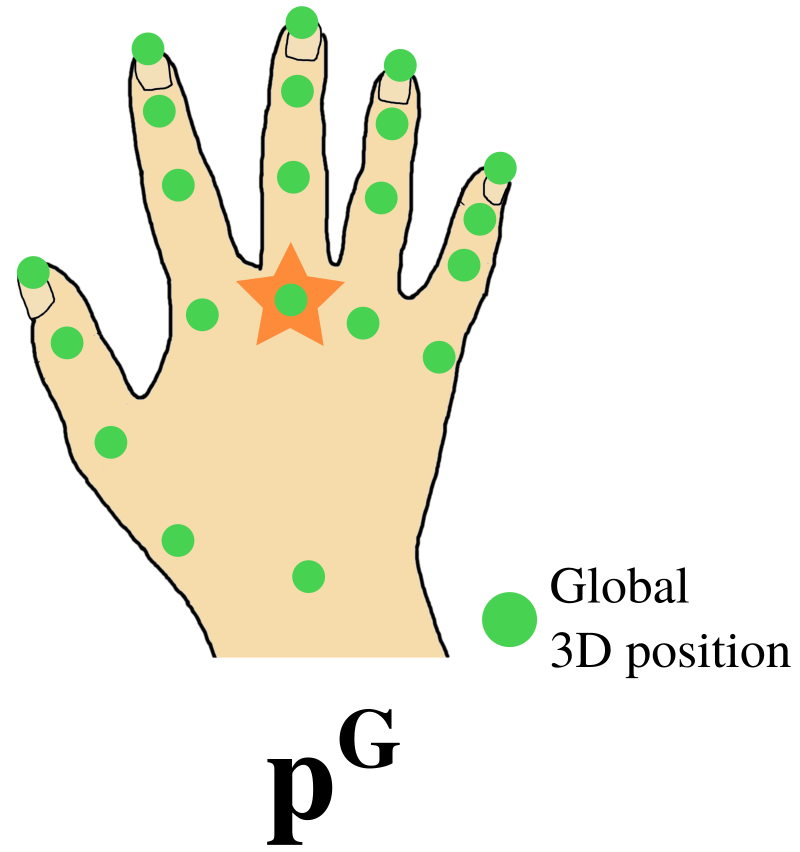}
        \caption{Global 3D positions}
        \label{fig:hand-model-3dpos}
    \end{subfigure}
    \begin{subfigure}[t]{0.45\columnwidth}
        \includegraphics[width=\textwidth]{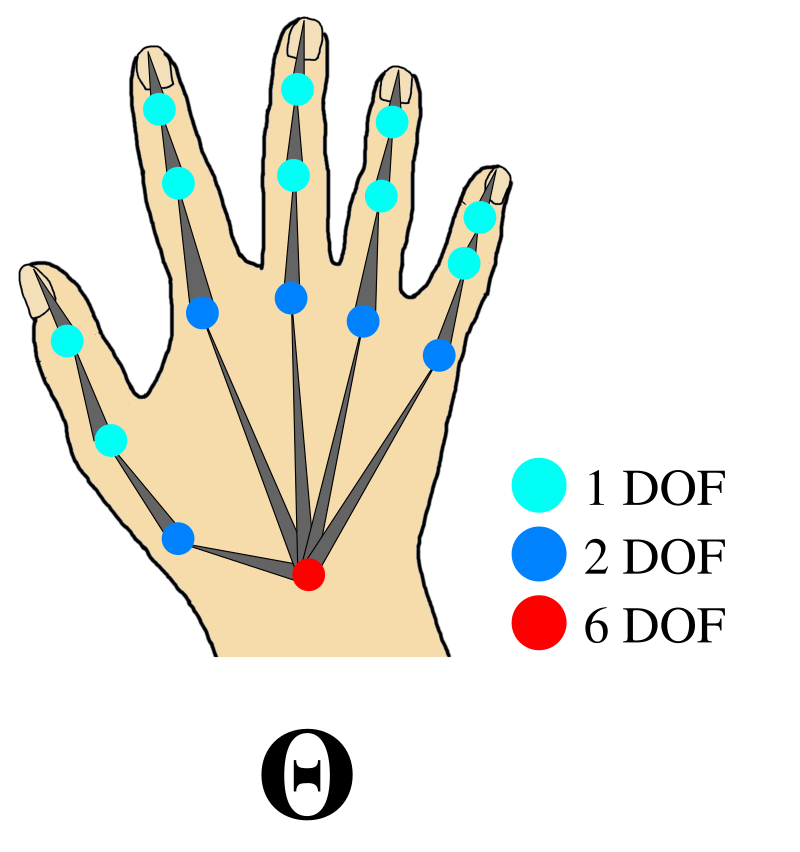}
        \caption{Kinematic skeleton}
        \label{fig:hand-model-skeleton}
    \end{subfigure}
	\caption{We use two different, but consistent, representations to model the hands.
    Our 3D joint regression step outputs $J=21$ global 3D joint locations, shown in (a) in green, which are later used to estimate the joint angles of a kinematic skeleton hand model, shown in (b).
    The orange star depicts the joint used as a hand root.}
	\label{fig:hand-model}
\end{figure}

To ensure a consistent representation for both joint locations (predicted by the CNNs) and joint angles (optimized during tracking), we use a kinematic skeleton.
As shown in Figure~\ref{fig:hand-model}, we model the hand using a hierarchy of bones (gray lines) and joints (circles).
The 3D joint locations are used as constraints in a kinematic pose tracking step that estimates temporally smooth joint angles of a kinematic skeleton.
In our implementation, we use a kinematic skeleton with 26 degrees of freedom (DOF), which includes 6 for global translation and rotation, and 20 joint angles, stored in a vector $\bm{\Theta}$, as shown in Figure \ref{fig:hand-model-skeleton}.
To fit users with different hand shapes and sizes, we perform a quick calibration step to fix the length of the bones for different users.

\section{Single Frame 3D Pose Regression}
\label{sec:hand-pose}
The goal of 3D pose regression is to estimate the 3D joint locations of the hand at each frame of the \mbox{RGB-D} input.
To achieve this, we first create a \emph{colored depth map} $\mathcal{D}$, from the raw input produced by commodity \mbox{RGB-D} cameras (\eg, Intel RealSense SR300).
We define $\mathcal{D}$ as
\begin{equation}
	\mathcal{D} = \texttt{colormap}(\textbf{R},\textbf{G},\textbf{B},\textbf{Z}),
\end{equation}
where $\texttt{colormap}(\cdot)$ is a function, that depends on the camera calibration parameters, to map each pixel in the color image plane onto the depth map $\textbf{Z}$.
Computing $\mathcal{D}$ allows us to ignore camera-specific variations in extrinsic parameters.
We also downsample $\mathcal{D}$ to a resolution of 320$\times$240 to aid real-time performance.
We next describe our pose regression approach that is robust even in challenging cluttered scenes with notable (self-)occlusions of the hand.
As we show in the evaluation (\cref{sec:results-and-evaluation}), using a two step approach to first localize the hand in full-frame input and subsequently estimate 3D pose outperforms using a single CNN for both tasks.

\subsection{Hand Localization}
\label{sec:hand-localization}
The goal of the first part of pose regression is to localize the hand in challenging cluttered input frames resulting in a bounding box around the hand and 3D root location.
Given a colored depth map $\mathcal{D}$, we compute
 \begin{equation}
    \croppedinput = \texttt{imcrop}(\mathcal{D}, H_\text{R}),
 \end{equation}
where $H_\text{R}$ is a heatmap encoding the position probability of the 2D hand root and $\texttt{imcrop}(\cdot)$ is a function that crops the hand area of the input frame. 
In particular, we estimate $H_\text{R}$ using a CNN which we call \emph{HALNet} (HAnd Localization Net).
The $\texttt{imcrop}(\cdot)$ function picks the image-level heatmap maximum location $\phi(H_\text{R}) = (u, v)$ and uses the associated depth $z$ in $\mathcal{D}$ to compute a depth-dependent crop, the side length of which is inversely proportional to the depth and contains the hand.
Additionally, $\texttt{imcrop}(\cdot)$ also normalizes the depth component of the cropped image by subtracting $z$ from all pixels.

\emph{HALNet} uses an architecture derived from \emph{ResNet50}~\cite{he_cvpr2016} which has been shown to have a good balance between accuracy and computational cost~\cite{canziani2016analysis}.
We reduced the number of residual blocks to 10 to achieve real-time framerate while maintaining high accuracy.
We train this network using \emph{SynthHands}, a new photorealistic dataset with ample variance across many dimensions such as hand pose, skin color, objects, hand-object interaction and shading details.
See Sections \ref{sec:SynthHands} and \ref{sec:training}, and the supplementary document for training and architecture details.

\parahead{Post Processing}
To make the root maximum location robust over time, we add an additional step to prevent outliers from affecting 3D joint location estimates.
We maintain a history of maxima locations and label them as \emph{confident} or \emph{uncertain} based on the following criterion.
If the likelihood value of the heatmap maximum at a frame $t$ is $< 0.1$ and it occurs at $>$ 30 pixels from the previous maximum then it is marked as uncertain.
If a maximum location is uncertain, we update it as
\begin{eqnarray}
\phi_t = \phi_{t-1} + \delta^k \, \frac{\phi_{c-1} - \phi_{c-2}}{||\phi_{c-1} - \phi_{c-2}||},
\end{eqnarray}
where $\phi_t = \phi(H_\text{R}^t)$ is the updated 2D maximum location at the frame $t$, $\phi_{c-1}$ is the last \emph{confident} maximum location, $k$ is the number of frames elapsed since the last confident maximum, and $\delta$ is a decay factor to progressively downweight uncertain maxima.
We empirically set $\delta = 0.98$ and use this value in all our results.

\subsection{3D Joint Regression}
\label{sec:3d-joint-regression}
Starting from a cropped and normalized input $\croppedinput$ that contains a hand, potentially partially occluded, our goal is to regress the global 3D hand joint position vector $\mathbf{p^G} \in \mathbb{R}^{3 \times J}$.
We use a CNN, referred to as \emph{JORNet} (JOint Regression Net), to predict per-joint 3D root-relative positions $\mathbf{p^L} \in \mathbb{R}^{3 \times J}$ in $\croppedinput$.
Additionally, \emph{JORNet} also regresses per-joint 2D position likelihood heatmaps $\mathbf{H} = \{H_j\}_{j=1}^{J}$, which will be used to regularize the predicted 3D joint positions in a later step.
We obtain global 3D joint positions $\mathbf{p}^{\mathbf{G}}_j = \mathbf{p}^{\mathbf{L}}_j + \mathbf{r}$, where $\mathbf{r}$ is the global position of the hand center (or a point on an occluder) obtained by backprojecting the 2.5D hand root position ($u,v,z$) to 3D.
\emph{JORNet} uses the same architecture as \emph{HALNet} and is trained with the same data.
See Sections \ref{sec:SynthHands} and \ref{sec:training} for training details, and the supplementary document for architecture details. 

\begin{figure}
	\centering
    
    \begin{subfigure}[t]{0.24\columnwidth}
    	\includegraphics[width=\textwidth]{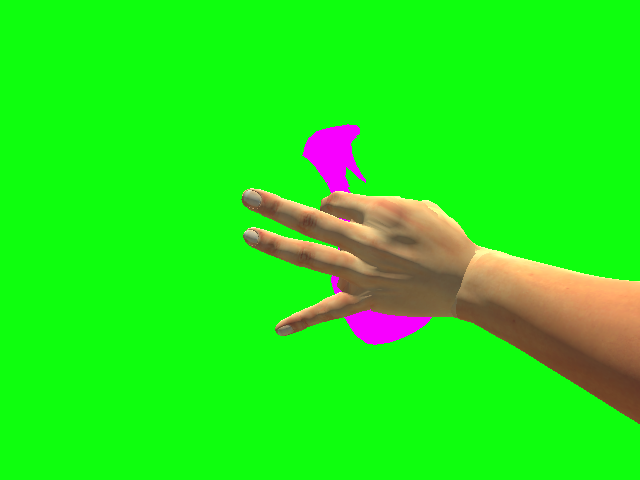}
    \end{subfigure}
    \begin{subfigure}[t]{0.24\columnwidth}
    	\includegraphics[width=\textwidth]{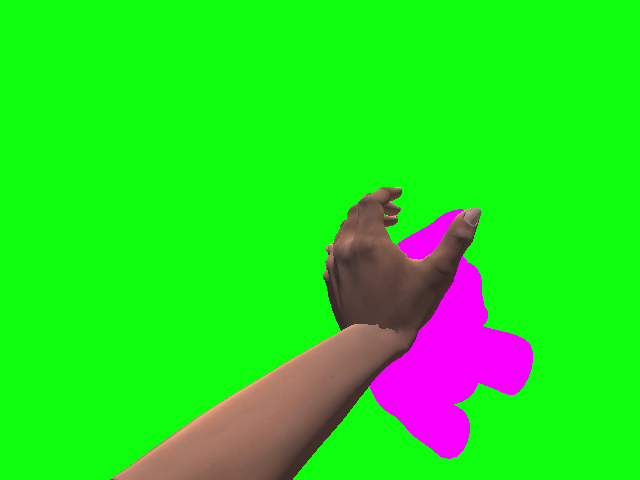}
    \end{subfigure}
    \begin{subfigure}[t]{0.24\columnwidth}
    	\includegraphics[width=\textwidth]{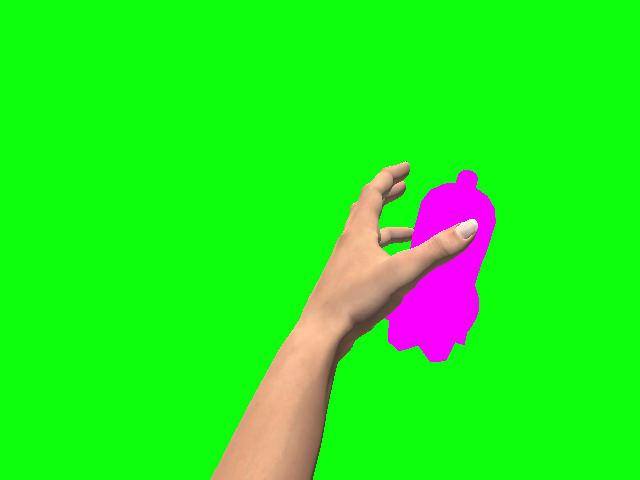}
    \end{subfigure}
    \begin{subfigure}[t]{0.24\columnwidth}
    	\includegraphics[width=\textwidth]{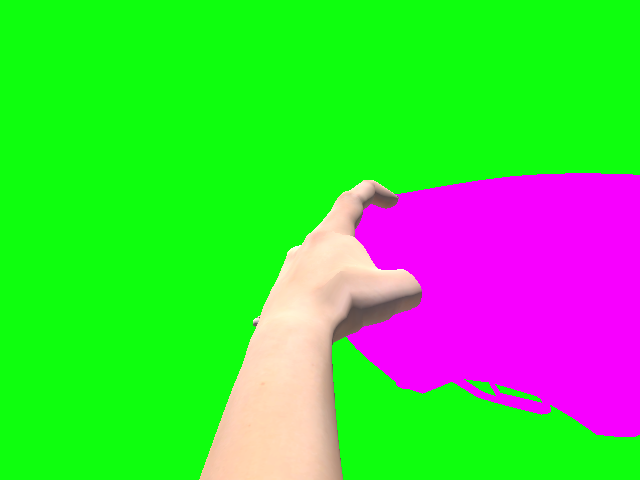}
    \end{subfigure}   
    
    \vspace{0.08cm}
    \begin{subfigure}[t]{0.24\columnwidth}
    	\includegraphics[width=\textwidth]{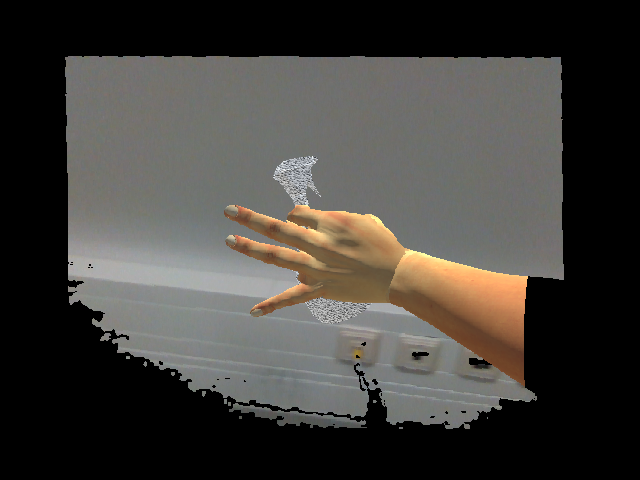}
    \end{subfigure}
    \begin{subfigure}[t]{0.24\columnwidth}
    	\includegraphics[width=\textwidth]{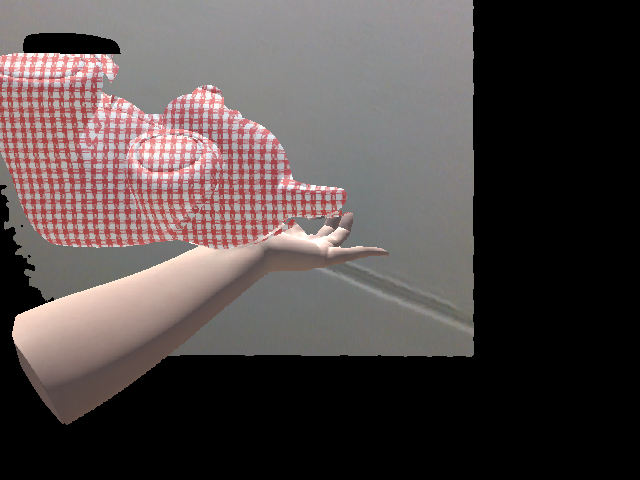}
    \end{subfigure}
    \begin{subfigure}[t]{0.24\columnwidth}
    	\includegraphics[width=\textwidth]{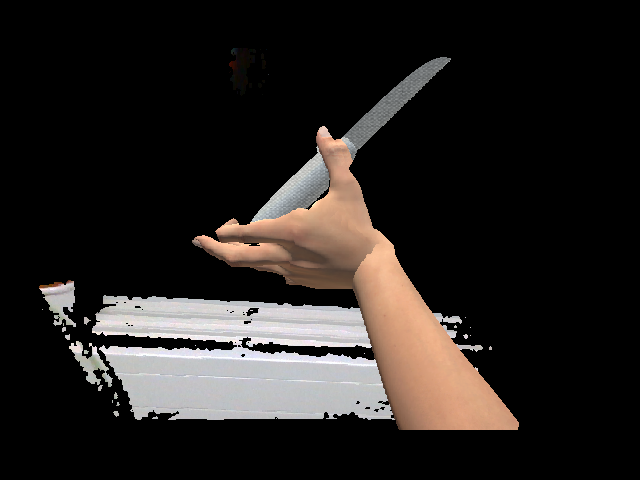}
    \end{subfigure}
    \begin{subfigure}[t]{0.24\columnwidth}
    	\includegraphics[width=\textwidth]{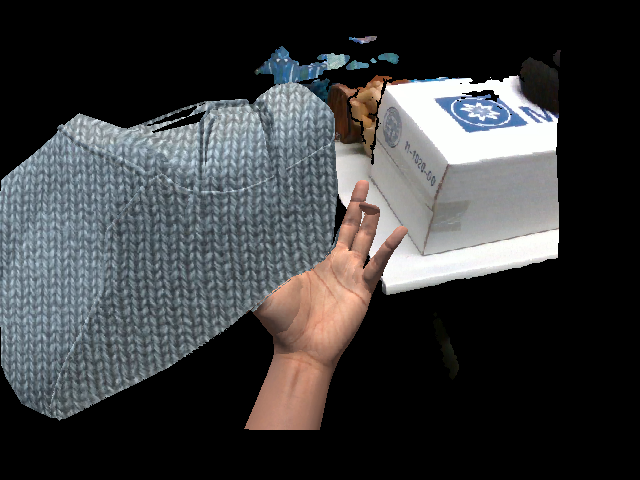}
    \end{subfigure}  
    
	\caption{Our \textit{SynthHands} dataset is created by posing a photorealistic hand model with real hand motion data.
    Virtual objects are incorporated into the 3D scenario.
    To allow data augmentation, we output object foreground and scene background appearance as a constant plain color (top row), which are composed with shading details and randomized textures in a postprocessing step (bottom row).}
	\label{fig:SynthHands}
\end{figure}

\begin{figure}
	\centering
	\begin{subfigure}[t]{0.975\columnwidth}
        \includegraphics[width=\textwidth]{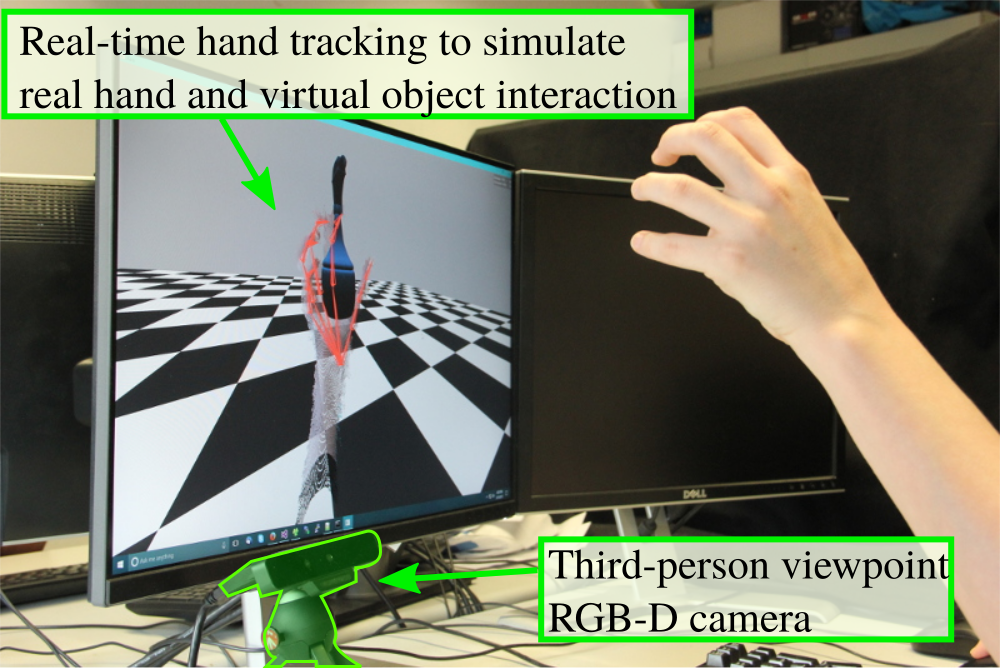}
    \end{subfigure}
    
    \vspace{0.05cm}
    \begin{subfigure}[t]{0.32\columnwidth}
        \includegraphics[trim = {0 2cm 0 2cm}, clip, width=\textwidth]{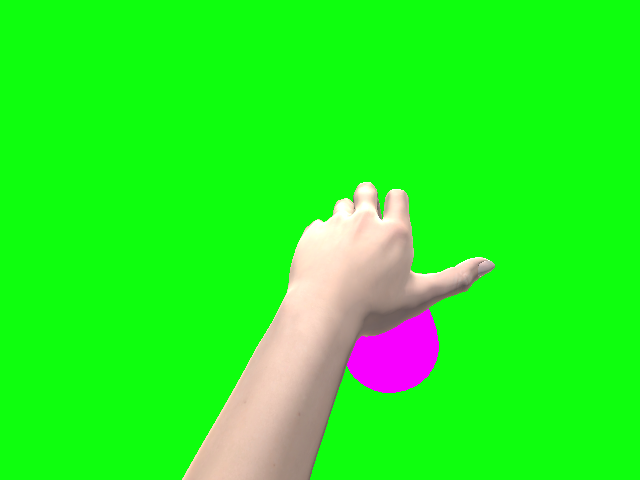}
    \end{subfigure}
    \begin{subfigure}[t]{0.32\columnwidth}
        \includegraphics[trim = {0 2cm 0 2cm}, clip, width=\textwidth]{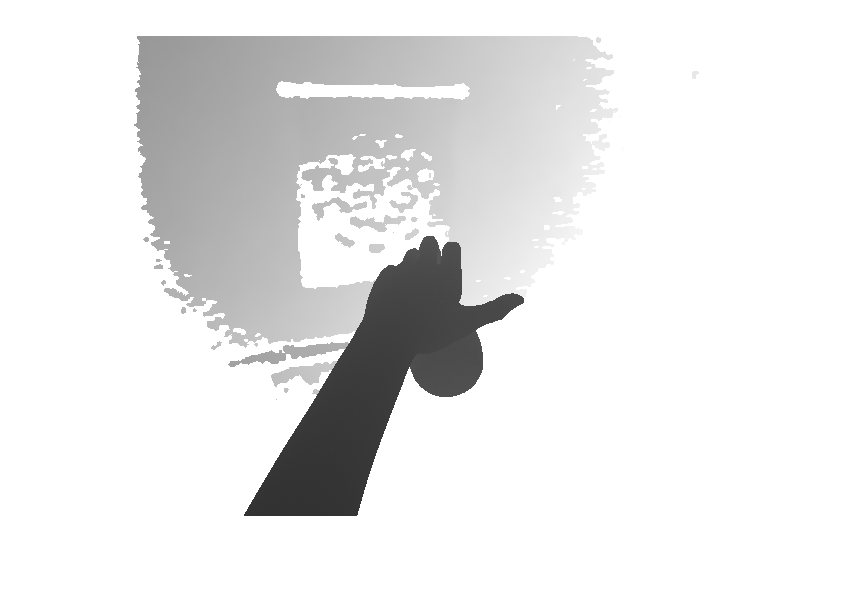}
    \end{subfigure}
    \begin{subfigure}[t]{0.32\columnwidth}
        \includegraphics[trim = {0 2cm 0 2cm}, clip, width=\textwidth]{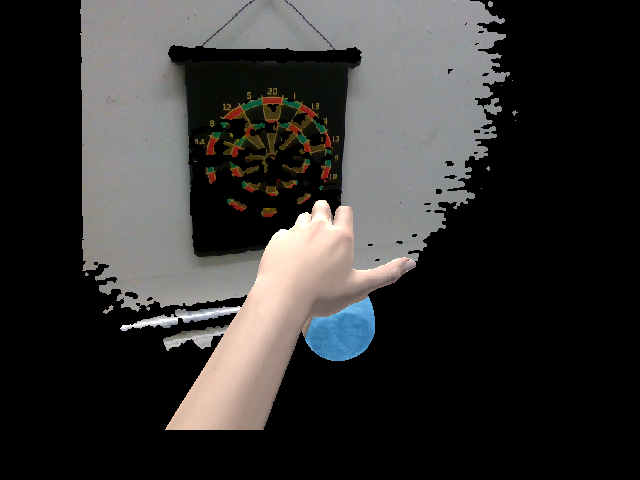}
    \end{subfigure}
    \caption{Our \textit{SynthHands} dataset has accurate annotated data of a hand interacting with objects.
    We use a merged reality framework to track a real hand, where all joint positions are annotated, interacting with a virtual object (top).
    Synthetic images are rendered with chroma key-ready colors, enabling data augmentation by composing the rendered hand with varying object texture and real cluttered backgrounds (bottom).}
	\label{fig:hand-object-interaction}
\end{figure}
\subsection{\textit{\textbf{SynthHands}} Dataset}
\label{sec:SynthHands}
Supervised learning methods, including CNNs, require large amounts of training data in order to learn all the variation exhibited in real hand motion.
Fully annotated real data would be ideal for this purpose but it is time consuming to manually annotate data and annotation quality may not always be good~\cite{oberweger_cvpr2016}.
To circumvent this problem, existing methods~\cite{rogez_eccv2014workshop,rogez_cvpr2015} have used synthetic data.
Despite the advances made, existing datasets are constrained in a number of ways: they typically show unnatural mid-air motions, no complex hand-object interactions, and do not model realistic background clutter and noise.

We propose a new dataset, \emph{SynthHands}, that combines real captured hand motion (retargeted to a virtual hand model) with natural backgrounds and virtual objects to sample all important dimensions of variability at previously unseen granularity.
It captures the variations in natural hand motion such as pose, skin color, shape, texture, background clutter, camera viewpoint, and hand-object interactions.
We now highlight some of the unique features of this dataset that make it ideal for supervised training of learning-based methods.

\parahead{Natural Hand Motions}
Instead of using static hand poses~\cite{rogez_cvpr2015}, we captured real, non-occluded, hand motion in mid-air from a third-person viewpoint, with a state-of-the-art real-time markerless tracker~\cite{sridhar_cvpr2015}.
These motions were subsequently re-targeted onto a photorealistic synthetic hand rigged by an artist.
Because we pose the synthetic hand using the captured hand motion, it mimics real hand motions and increases dataset realism.

\parahead{Hand Shape and Color}
Hand shape and skin color exhibit large variation across users.
To simulate real world diversity, \textit{SynthHands} contains skin textures randomly sampled from 12 different skin tones.
We also sample variation in other anatomical features (\eg, male hands are typically bigger and may contain more hair) in the data.
Finally, we model hand shape variation by randomly applying a scaling parameter $\beta \in [0.8, 1.2]$ along each dimension of a default hand mesh.

\parahead{Egocentric Viewpoint}
Synthetic data has the unique advantage that we can render from arbitrary camera viewpoints.
In order to support difficult egocentric views, we setup 5 virtual cameras that mimic different egocentric perspectives.
The virtual cameras generate \mbox{RGB-D} images from this perspective while also simulating sensor noise and camera calibration parameters.

\parahead{Hand-Object Interactions}
We realistically simulate hand-object interactions by using a merged reality approach to track real hand motion interacting with virtual objects.
We achieve this by leveraging the real-time capability of existing hand tracking solutions~\cite{sridhar_cvpr2015} to show the user's hand interacting with a virtual on-screen object.
Users perform motions such as object grasping and manipulation, thus simulating real hand-object interactions (see Figure \ref{fig:hand-object-interaction}).

\parahead{Object Shape and Appearance}
\textit{SynthHands} contains interactions with a total of 7 different virtual objects in various locations, rotations and scale configurations.
To enable augmentation of the object appearance to increase dataset variance, we render the object albedo (\ie, pink in Figure \ref{fig:SynthHands}) and shading layers separately.
We use chroma keying to replace the pink object albedo with a texture randomly sampled from a set of 145 textures and combining it with the shading image. Figure~\ref{fig:SynthHands} shows some examples of the data before and after augmentation. Importantly, note that \textit{SynthHands} does not contain 3D scans of the real test objects nor 3D models of similar objects used for evaluation in Section \ref{sec:results-and-evaluation}. This demonstrates that our approach generalizes to unseen objects.

\parahead{Real Backgrounds}
Finally, we simulate cluttered scenes and backgrounds by compositing the synthesized hand-object images with real \mbox{RGB-D} captures of real backgrounds, including everyday desktop scenarios, offices, corridors and kitchens.
We use chroma keying to replace the default background (green in Figure~\ref{fig:SynthHands}) with the captured backgrounds.

Our data generation framework is built using the Unity Game Engine and uses a rigged hand model distributed by Leap Motion \cite{leap_motion}.
In total, \textit{SynthHands} contains roughly 220,000 \mbox{RGB-D} images exhibiting large variation seen in natural hands and interactions.
Please see the supplementary document for more information and example images.

\subsection{Training}
\label{sec:training}
Both \emph{HALNet} and \emph{JORNet} are trained on the \textit{SynthHands} dataset using the Caffe framework~\cite{jia2014caffe}, and the AdaDelta solver with a momentum of 0.9 and weight decay factor of 0.0005.
The learning rate is tapered down from 0.05 to 0.000025 during the course of the training.
For training \emph{JORNet}, we used the ground truth $(u, v)$ and $z$ of the hand root to create the normalized crop input.
To improve robustness, we also add uniform noise ($\in [-25,25]\text{~mm}$) to the backprojected 3D root position in the \emph{SynthHands} dataset.
We trained \emph{HALNet} for 45,000 iterations and \emph{JORNet} for 60,000 iterations.
The final networks were chosen as the ones with the lowest loss values.
The layers in our networks that are similar to \emph{ResNet50} are initialized with weights of the original \emph{ResNet50} architecture trained on ImageNet~\cite{ILSVRC15}.
For the other layers, we initialize the weights randomly.
For details of the loss weights used and the taper scheme, please see the supplementary document.
\section{Hand Pose Optimization} \label{sec:hand-pose-opt}
The estimated per-frame global 3D joint positions $\mathbf{p}^\mathbf{G}$ are not guaranteed to be temporally smooth nor do they have consistent inter-joint distances (\ie, bone lengths) over time.
We mitigate this by fitting a kinematic skeleton parameterized by joint angles $\bm{\Theta}$, shown in Figure \ref{fig:hand-model-skeleton}, to the regressed 3D joint positions.
Additionally, we refine the fitting by leveraging the 2D heatmap output from \emph{JORNet} as an additional contraint and regularize it using joint limit and smoothness constraints.
In particular, we seek to minimize 
\begin{eqnarray}
\mathcal{E}(\bm{\Theta})= E_{\text{data}}(\bm{\Theta},\mathbf{p^G},\mathbf{H}) + E_{\text{reg}}(\bm{\Theta}), 
\end{eqnarray}
where $E_{\text{data}}$ is the data term that incorporates both the 3D positions and 2D heatmaps 
\begin{align}
E_{\text{data}}(\bm{\Theta},\mathbf{p^G},\mathbf{H})&= w_{\text{p3}}E_{\text{pos3D}}(\bm{\Theta},\mathbf{p^G}) + w_{\text{p2}} E_{\text{pos2D}}(\bm{\Theta},\mathbf{H}).
\end{align}
The first term $E_{\text{pos3D}}$ minimizes the 3D distance between each predicted joint location $\mathbf{p}^{\mathbf{G}}_j$ and its corresponding position $\mathcal{M}(\bm{\Theta})_j$ in the kinematic skeleton set to pose $\bm{\Theta}$
\begin{equation}
E_{\text{pos3D}}(\bm{\Theta}) = \sum_{j = 1}^{J} || \mathcal{M}(\bm{\Theta})_j - \mathbf{p}^{\mathbf{G}}_j||_2^2.
\end{equation}
The second data term, $E_{\text{pos2D}}$, minimizes the 2D distance between each joint heatmap maximum $\phi(H_j)$ and the projected 2D location of the corresponding joint in the kinematic skeleton
\begin{equation}
E_{\text{pos2D}}(\bm{\Theta}) = \sum_{j = 1}^{J} || \pi(\mathcal{M}(\bm{\Theta})_j) - \phi(H_j))||_2^2,
\end{equation}
where $\pi$ projects the joint onto the image plane.
We empirically tuned the weights for the different terms as: $w_{p3} = 0.01$ and $w_{p2} = 5\times10^{-7}$.

We regularize the data terms by enforcing joint limits and temporal smoothness constraints
\begin{align}
E_{\text{reg}}(\bm{\Theta}) &= w_{\text{l}} E_{\text{lim}}(\bm{\Theta}) + w_{\text{t}} E_{\text{temp}}(\bm{\Theta})
\end{align}
where 
\begin{equation}
E_{\text{lim}}(\bm{\Theta}) = \sum_{\theta_i \in \bm{\Theta}}
  \begin{cases}
    0 &, \text{if } \theta_i^{l} \le \theta_i \le \theta_i^u \\
    (\theta_i  -  \theta_i^l)^2 &, \text{if }  \theta_i  < \theta_i^l \\
    (\theta_i^u  - \theta_i)^2 &, \text{if }   \theta_i > \theta_i^u \\
  \end{cases}
\end{equation}
is a soft prior to enforce biomechanical pose plausibility, with $\bm{\Theta}^l, \bm{\Theta}^u$ being the lower and upper joint angle limits, respectively, and
\begin{equation}
E_{\text{temp}}(\bm{\Theta}) = ||\nabla \bm{\Theta} - \nabla \bm{\Theta}^{(t-1)}||_2^2
\end{equation}
enforces constant velocity to prevent dramatic pose changes.
We empirically chose weights for the regularizers as: $w_l = 0.03$ and $w_t = 10^{-3}$.
We optimize our objective using 20 iterations of conditioned gradient descent.
\section{Results and Evaluation}
\label{sec:results-and-evaluation}
We conducted several experiments to evaluate our method and different components of it.
To facilitate evaluation, we captured a new benchmark dataset \RealBenchmarkName~consisting of 3190 frames of natural hand interactions with objects in real cluttered scenes, moving egocentric viewpoints, complex hand-object interactions, and natural lighting.
Of these, we manually annotated 1485 frames using an annotation tool to mark 2D and 3D fingertip positions, a common approach used in free hand tracking~\cite{ballan_eccv2012,sridhar_iccv2013}.
In total we gathered 4 sequences (\texttt{Rotunda}, \texttt{Desk}, \texttt{Kitchen}, \texttt{Fruits}) featuring 4 different users (2 female), skin color variation, background variation, different objects, and camera motion.
Note that the objects in \RealBenchmarkName~are distinct from the objects in the \emph{SynthHands} training data to show the ability of our approach to generalize.
In addition, to enable evaluation of the different components of our method, we also held out a test set consisting of 5120 fully annotated frames from the \textit{SynthHands} dataset.

\parahead{Component Evaluation}
We first analyze the performance of \emph{HALNet} and \emph{JORNet} on the synthetic test set.
The main goal of \emph{HALNet} is to accurately localize the 2D position of the root (which either lies on the hand or on an occluder in front) accurately.
We thus use 2D Euclidean pixel error between the ground truth root position and the predicted position as the evaluation metric.
On average, \emph{HALNet} produces an error of \textbf{2.2~px} with a standard deviation of 1.5~px on the test set.
This low average error ensures that we always obtain reliable crops for \emph{JORNet}.

\begin{figure}
  \centering
  \begin{subfigure}[t]{0.48\columnwidth}
    \includegraphics[width=\columnwidth]{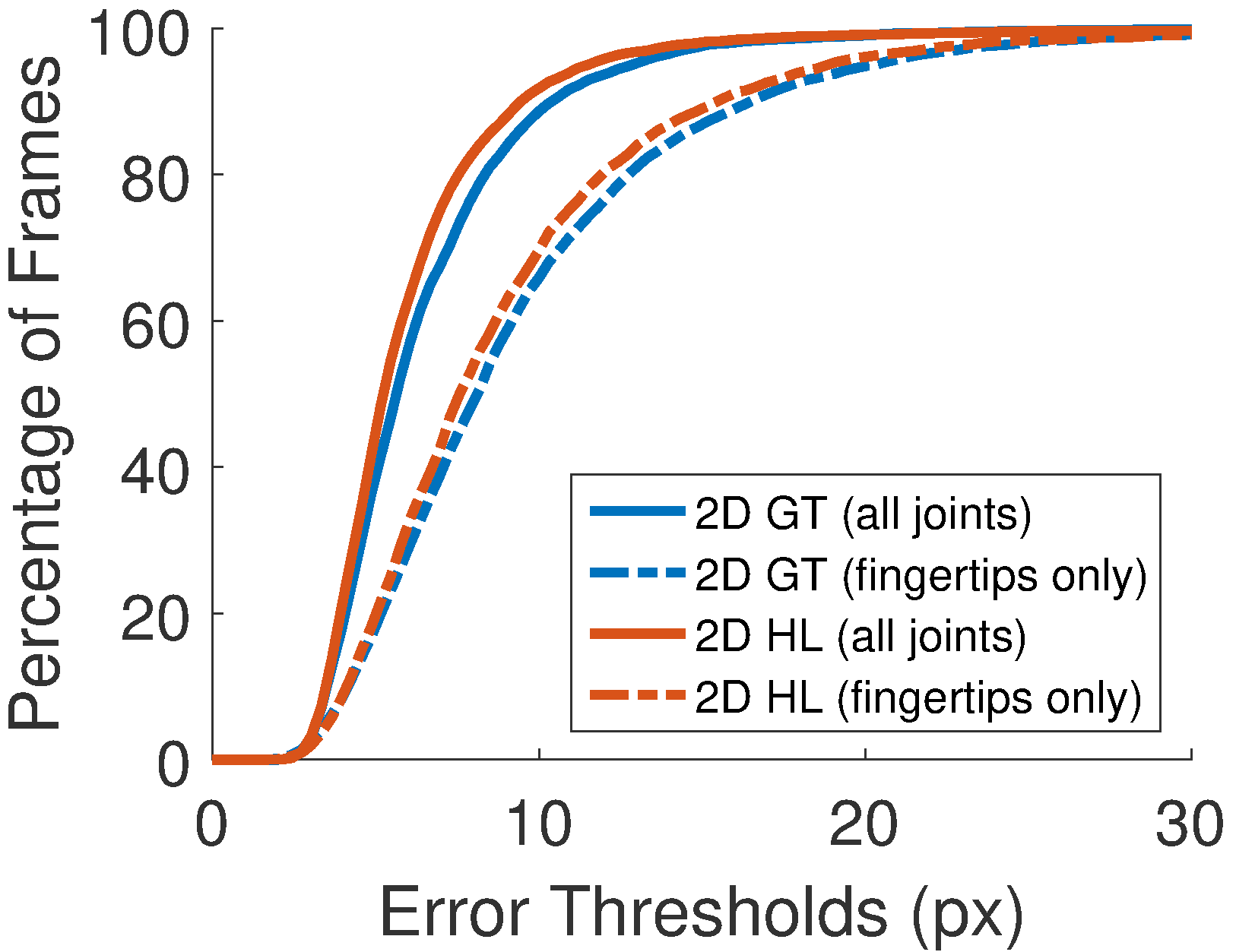}
  \end{subfigure}
  \begin{subfigure}[t]{0.48\columnwidth}
    \includegraphics[width=\columnwidth]{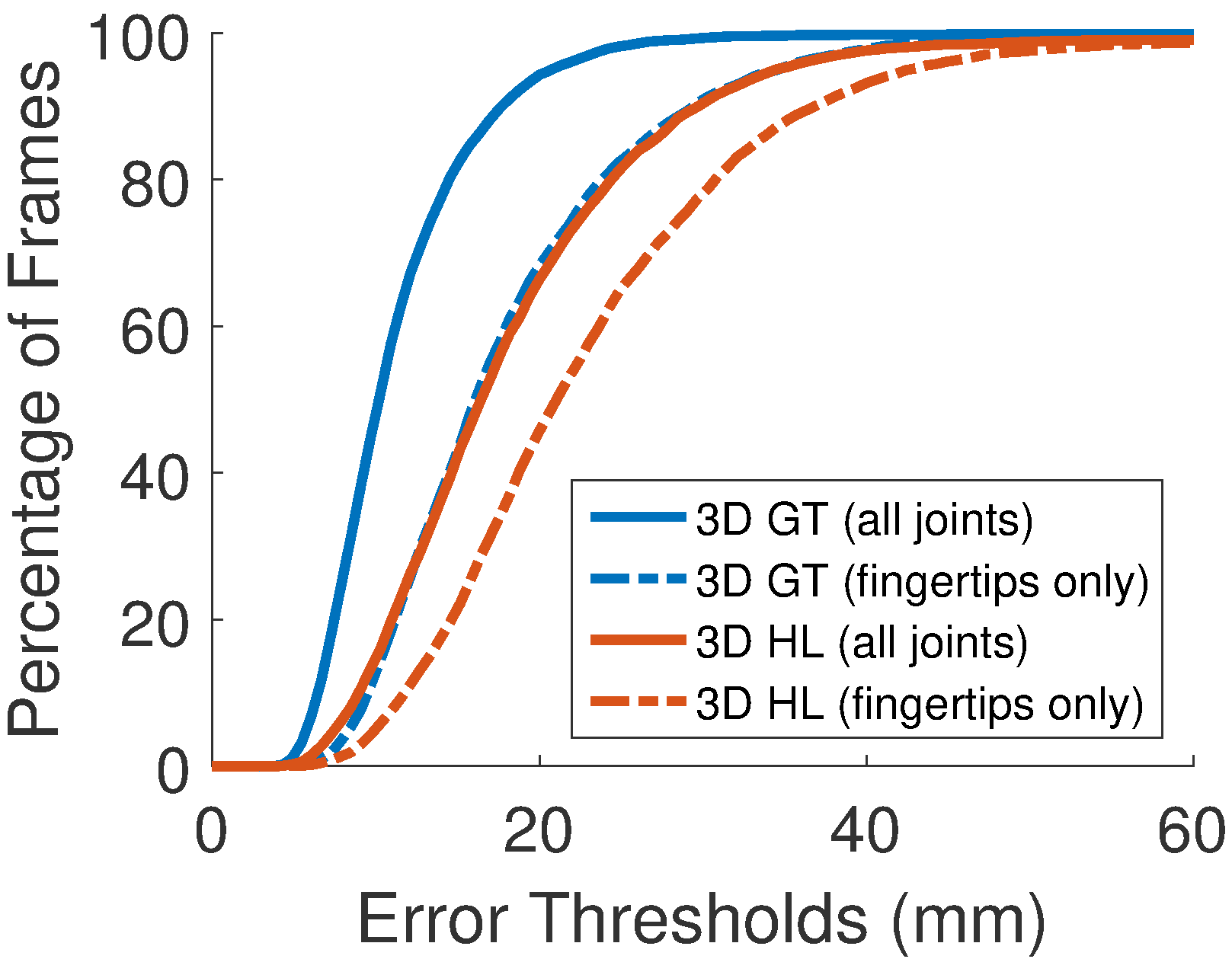}
  \end{subfigure}
  \caption{Comparison of 2D (left) and 3D (right) error of the joint position estimates of \emph{JORNet}.
    \emph{JORNet} was initialized with either the ground truth (GT, blue) or with the proposed hand localization step (HL, orange).
    We observe that HL initialization does not substantially reduce the performance of \emph{JORNet}.
    As expected, fingertips-only errors (dashed lines) are higher than the errors for all joints.
  }
  \label{fig:jornet-quant}
\end{figure}

To evaluate \emph{JORNet}, we use the 3D Euclidean distance between ground truth joint positions (of all hand joints) and the predicted position as the error metric.
For comparison, we also report the errors for only the 3D fingertip positions which are a stricter measure of performance.
Since the output of \emph{JORNet} is dependent on the crop estimated in the hand localization step, we evaluate two conditions: (1) using ground truth crops, (2) using crops from the hand localization step.
This helps evaluate how hand localization affects the final joint positions.
Figure~\ref{fig:jornet-quant} shows the percentage of the test set that produces a certain 2D or 3D error for all joints and fingertips only.
For 3D error, we see that using ground truth (GT) crops is better than using the crops from the hand localization (HL).
The difference is not substantial which shows that the hand localization step does not lead to catastrophic failures of \emph{JORNet}.
For 2D error, however, we observe that \emph{JORNet} initialized with HL results in marginally better accuracy.
We hypothesize that this is because \emph{JORNet} is trained on noisy root positions (\cref{sec:training}) while the ground truth lacks any such noise.

\begin{figure}
  \centering
  \includegraphics[width=0.98\columnwidth]{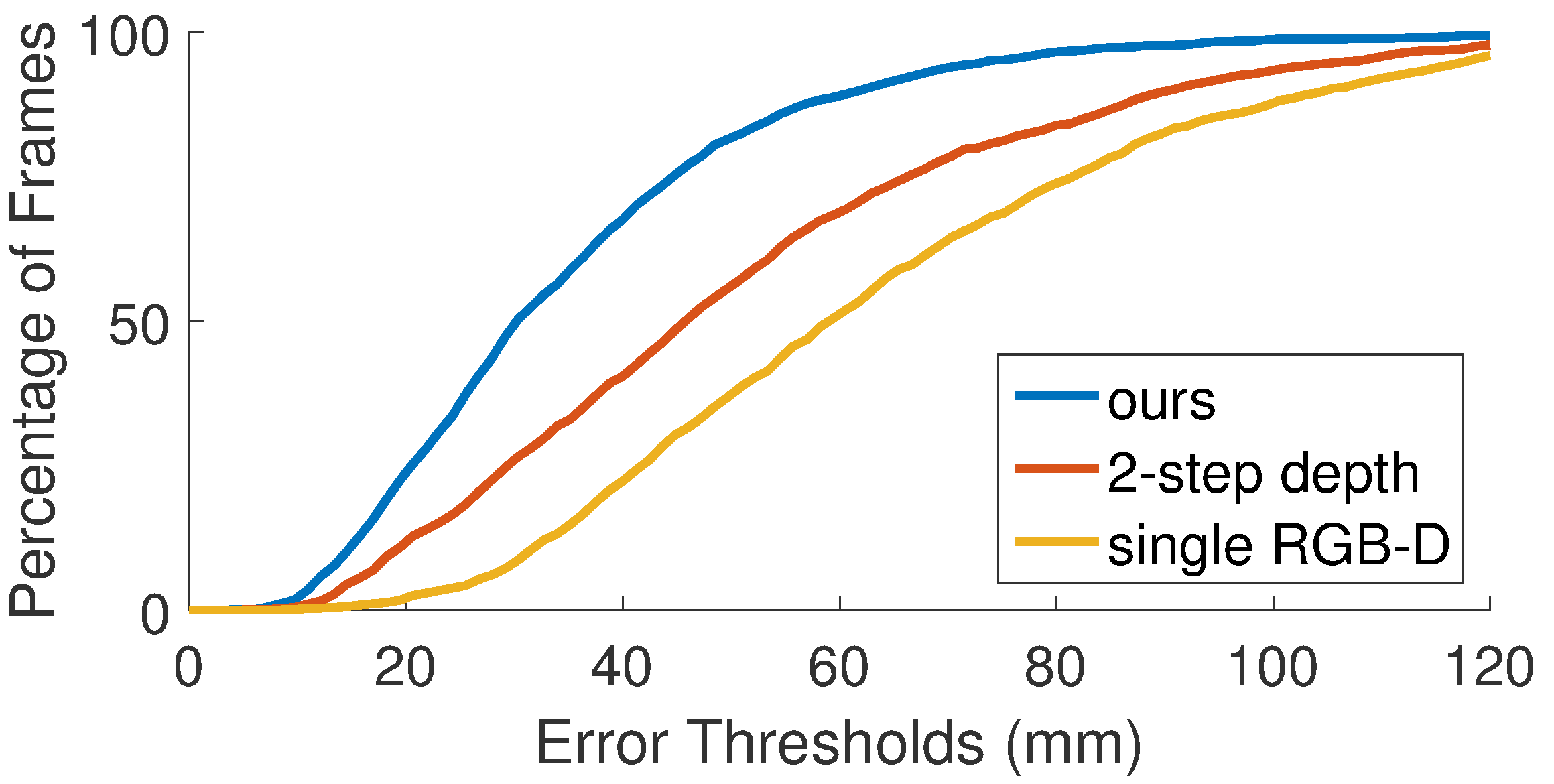}
  \caption{Comparison of our two-step RGB-D CNN architecture, the corresponding depth-only version and a single combined CNN which is trained to directly regress global 3D pose. Our proposed approach achieves the best performance on the real test sequences.}
  \label{fig:single-vs-two_rgbd-vs-depth}
\end{figure}

\parahead{CNN Structure Evaluation}
We now show that, on our real annotated benchmark \RealBenchmarkName, our approach that uses two subsequently applied CNNs is better than a single CNN to directly regress joint positions in cluttered scenes.
We trained a CNN with the same architecture as \emph{JORNet} but with the task of directly regressing 3D joint positions from full frame \mbox{RGB-D} images which often have large occlusions and scene clutter.
In Figure~\ref{fig:single-vs-two_rgbd-vs-depth}, we show the 3D fingertip error plot for this CNN (single \mbox{RGB-D}) which is worse that our two-step approach.
This shows that learning to directly regress 3D pose in cluttered scenes with occlusion is a harder task, which our approach simplifies by breaking it into two steps.
\begin{figure}
  \centering
  \includegraphics[width=0.95\columnwidth,height=5cm]{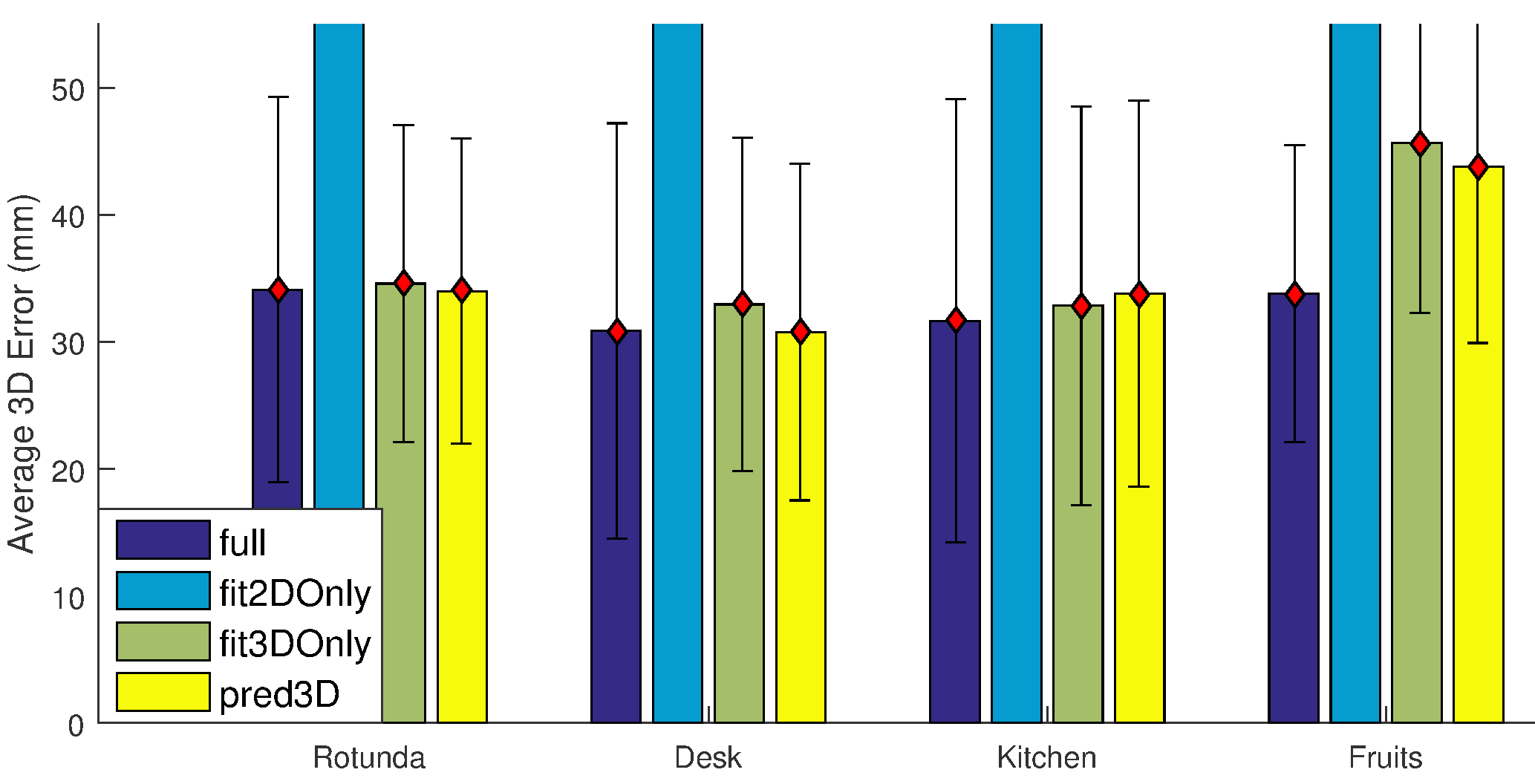}
  \caption{Ablative analysis of the proposed kinematic pose tracking on our real annotated dataset \RealBenchmarkName~(average fingertip error).
    Using only the 2D fitting energy leads to catastrophic tracking failure on all sequences.
    The version restricted to the 3D fitting term achieves a similar error as the raw 3D predictions while it ensures biomechanical plausibility and temporal smoothness.
    Our full formulation that combines 2D as well as 3D terms yields the lowest error.}
  \label{fig:kinematic-final}
  \vspace{-0.15cm}
\end{figure}

\begin{figure*}[!ht]
\centering
	\begin{tikzpicture}[every node/.style={inner sep=0,outer sep=0}]
 
  		\node[](image_label){\includegraphics[width=2\columnwidth,trim={0 0 0 1cm},clip]{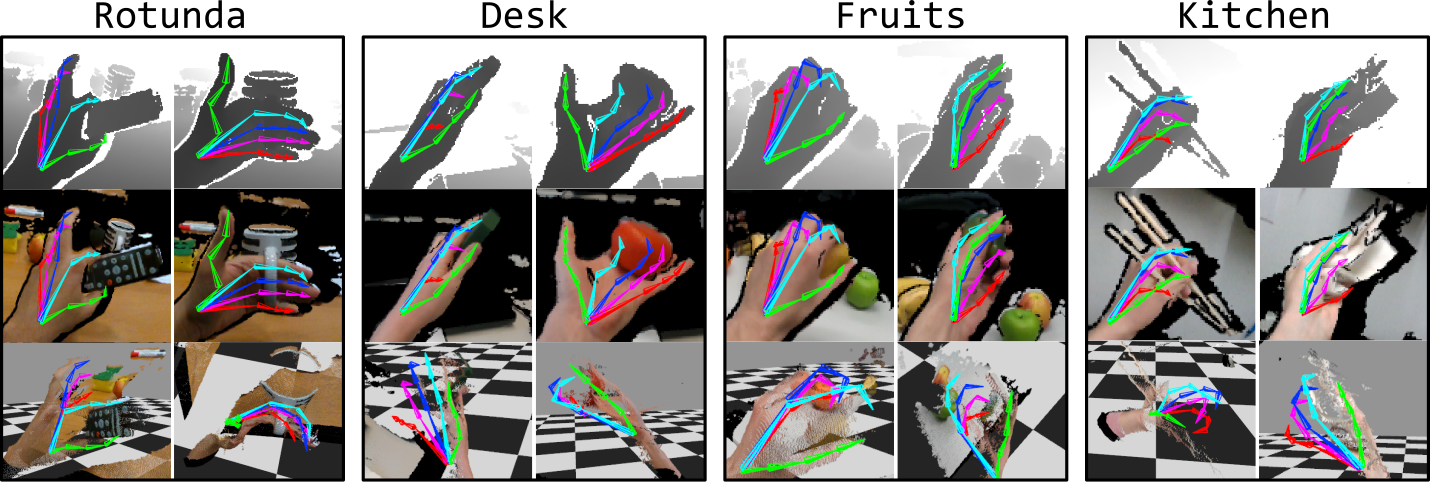}};
        \node[align = center, below left= 0cm and -2.85cm of image_label](rotunda_label){\texttt{Rotunda}};
        \node[align = center, right = 2.95cm of rotunda_label](desk_label){\texttt{Desk}};
        \node[align = center, right = 3.30cm of desk_label](fruits_label){\texttt{Fruits}};
        \node[align = center, right = 2.75cm of fruits_label](kitchen_label){\texttt{Kitchen}};
  	\end{tikzpicture}
  \caption{Qualitative results on our real annotated test sequences from the \RealBenchmarkName~benchmark dataset. The results overlayed on the input images and the corresponding 3D view from a virtual viewpoint (bottom row) show that our approach is able to handle complex object interactions, strong self-occlusions and a variety of users and backgrounds.}
  \label{fig:qualitative}
\end{figure*}
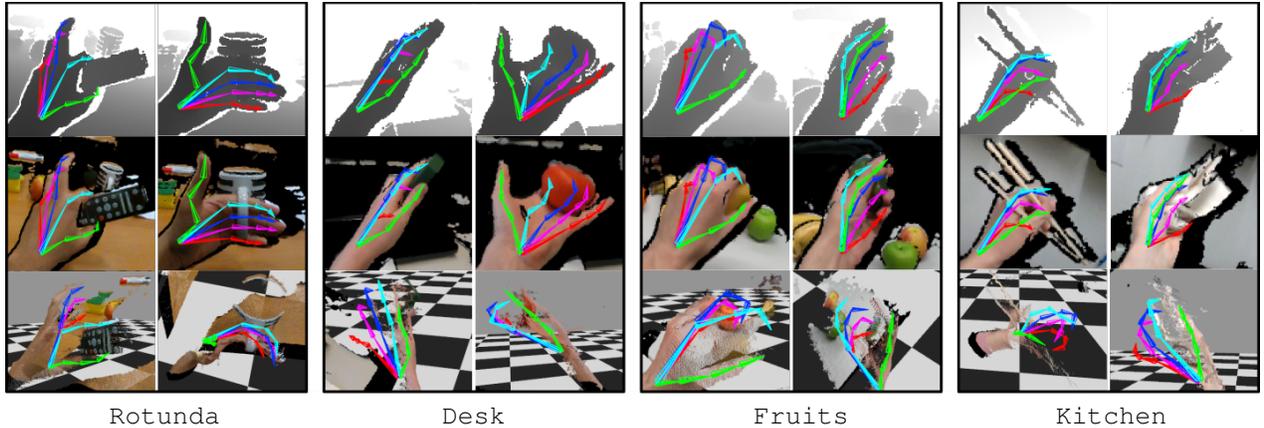

\parahead{Input Data Evaluation}
We next show, on our \RealBenchmarkName~dataset, that using both RGB and depth input (\mbox{RGB-D}) is superior to using only depth, even when using both our CNNs.
Figure~\ref{fig:single-vs-two_rgbd-vs-depth} compares the 3D fingertip error of a variant of our two-step approach trained with only depth data.
We hypothesize that additional color cues help our approach perform significantly better.

\parahead{Gain of Kinematic Model}
Figure~\ref{fig:kinematic-final} shows an ablative analysis of our energy terms as well as the effect of kinematic pose tracking on the final pose estimate. Because we enforce joint angle limits, temporal smoothness, and consistent bone lengths, our combined approach produces the lowest average error of \textbf{32.6~mm}.

We were unable to quantitatively evaluate on the only other existing \textit{egocentric} hand dataset~\cite{rogez_cvpr2015} due to a different sensor unsupported by our approach.
To aid qualitative comparison, we include similar reenacted scenes, background clutter, and hand motion in the supplemental document and video.

\begin{figure}
  \centering
  \includegraphics[width=\columnwidth]{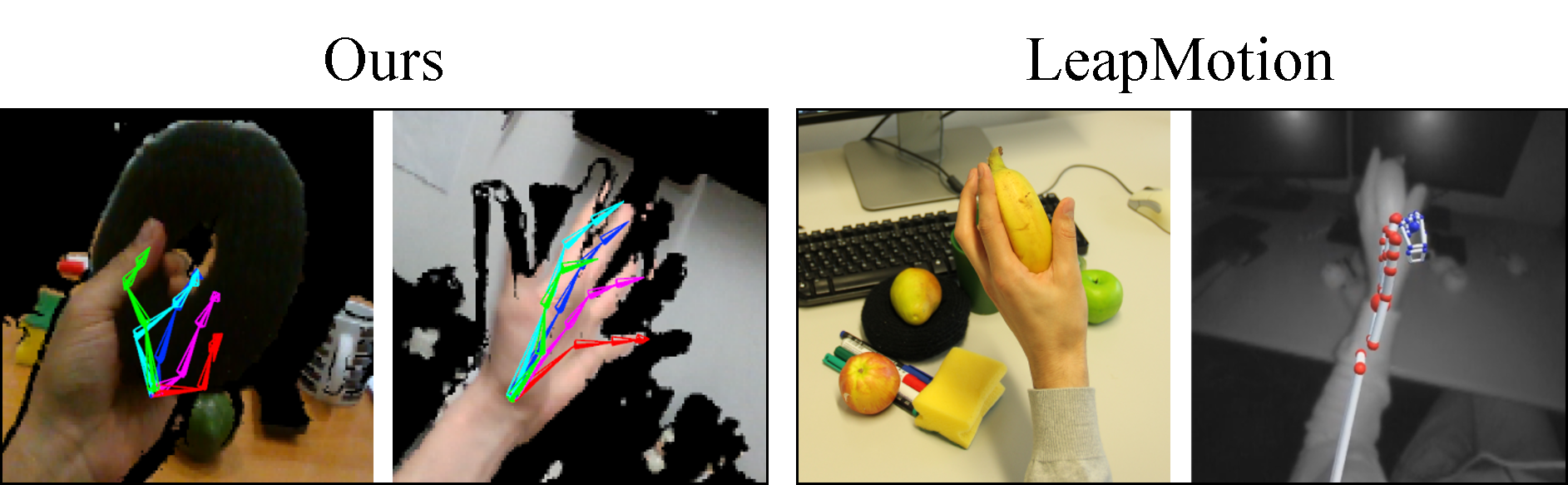}
  \caption{Fast motion that leads to misalignment in the colored depth image or failures in the hand localization step can lead to incorrect predictions (left two columns).
    LeapMotion Orion fails under large occlusions (right).}
  \label{fig:failures}
\end{figure}

\parahead{Qualitative Results}
Figure~\ref{fig:qualitative} shows qualitative results from our approach which works well for challenging real world scenes with clutter, hand-object interactions, and different hand shapes.
We also show that a commercial solution (LeapMotion Orion) does not work well under severe occlusions caused by objects, see Figure~\ref{fig:failures} right.
We refer to the supplemental document for results on how existing third person methods fail on \RealBenchmarkName~and how our approach in fact generalizes to third person views.

\parahead{Runtime Performance}
Our entire method runs in real-time on an Intel Xeon E5-2637 CPU (3.5~GHz) with an Nvidia Titan X (Pascal).
Hand localization takes 11~ms, 3D joint regression takes 6~ms, and kinematic pose tracking takes 1~ms.

\parahead{Limitations}
Our method works well even in challenging egocentric viewpoints and notable occlusions.
However, there are some failure cases which are shown in Figure~\ref{fig:failures}.
Please see the supplemental document for a more detailed discussion of failure cases.
We used large amounts of synthetic data for training our CNNs and simulated sensor noise for a specific camera preventing generalization.
In the future, we would like to explore the application of deep domain adaptation~\cite{ganin2014unsupervised} which offers a way to jointly make use of labeled \emph{synthetic} data together with unlabeled or partially labeled \emph{real} data.
\section{Conclusion}
We have presented a method for hand pose estimation in challenging first-person viewpoints with large occlusions and scene clutter.
Our method uses two CNNs to localize and estimate, in real time, the 3D joint locations of the hand.
A pose tracking energy further refines the pose by estimating the joint angles of a kinematic skeleton for temporal smoothness.
To train the CNNs, we presented \emph{SynthHands}, a new photorealistic dataset that uses a merged reality approach to capture natural hand interactions, hand shape, size and color variations, object occlusions, and background variations from egocentric viewpoints.
We also introduce a new benchmark dataset \RealBenchmarkName~that contains annotated sequences of challenging cluttered scenes as seen from egocentric viewpoints.
Quantitative and qualitative evaluation shows that our approach is capable of achieving low errors and consistent performance even under difficult occlusions, scene clutter, and background changes.

\parahead{Acknowledgements}
This work was supported by the ERC Starting Grant CapReal (335545).
Dan Casas was supported by a Marie Curie Individual Fellow, grant 707326.

\clearpage

\appendix
{
	\noindent
	\Large
	\textbf{Appendix}
}
\vspace{0.3cm}

This appendix includes additional experiments (Section~\ref{sec:eval}), details about our new synthetic dataset \emph{SynthHands} (Section~\ref{sec:synthhands}), and information about our custom CNN architecture and training procedure (Section~\ref{sec:CNN}).
We also refer to our supplemental video\footnote{http://handtracker.mpi-inf.mpg.de/projects/OccludedHands/} for more visual results.

\section{Evaluation} \label{sec:eval}
In this section, we show additional experiments and comparisons of our method.

\begin{figure*}[h]	
	\centering
	\begin{subfigure}[b]{0.03\textwidth}
		\includegraphics[trim={0 -2.0cm 0 0cm},clip,width=\textwidth]{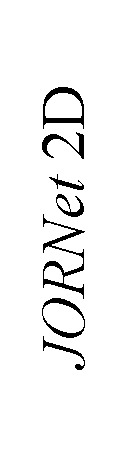}
	\end{subfigure}
	\hspace{0.05pt}
	\begin{subfigure}[b]{0.155\textwidth}
		\includegraphics[width=\textwidth]{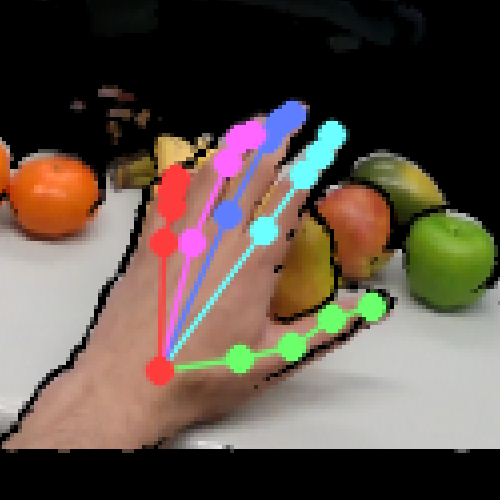}
	\end{subfigure}
	\begin{subfigure}[b]{0.155\textwidth}
		\includegraphics[width=\textwidth]{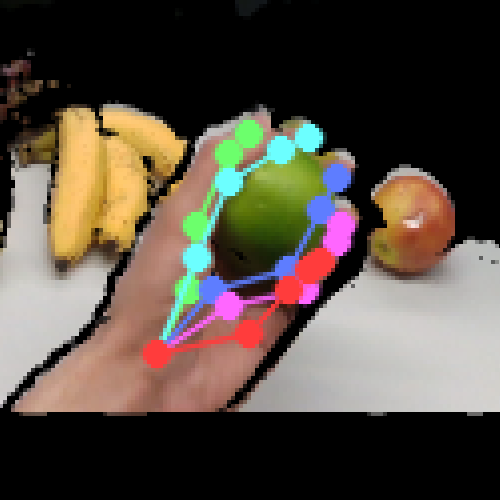}
	\end{subfigure}
	\begin{subfigure}[b]{0.155\textwidth}
		\includegraphics[width=\textwidth]{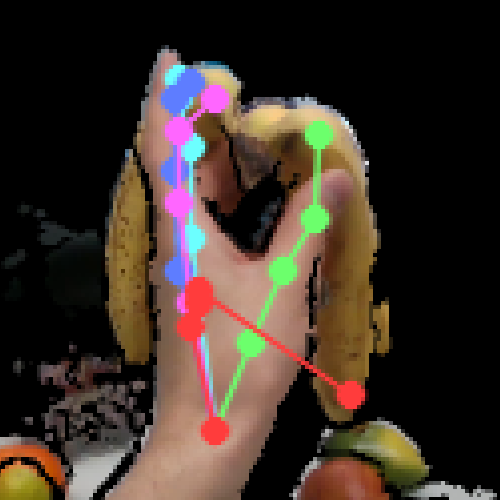}
	\end{subfigure}
	\begin{subfigure}[b]{0.155\textwidth}
		\includegraphics[width=\textwidth]{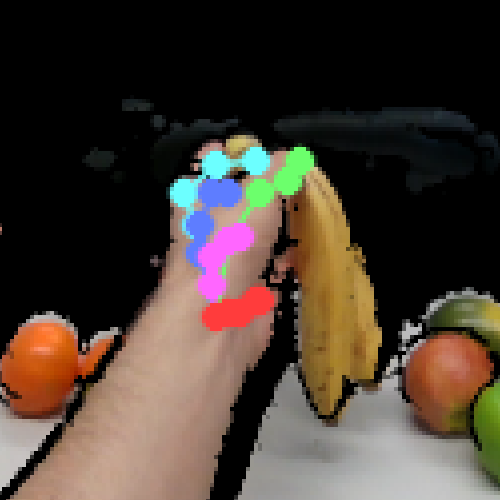}
	\end{subfigure}
	\begin{subfigure}[b]{0.155\textwidth}
		\includegraphics[width=\textwidth]{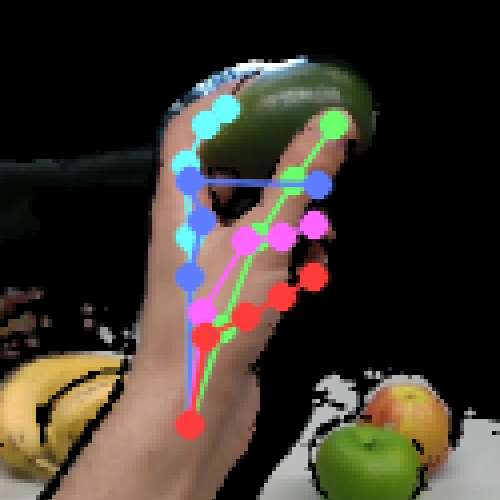}
	\end{subfigure}
	\begin{subfigure}[b]{0.155\textwidth}
		\includegraphics[width=\textwidth]{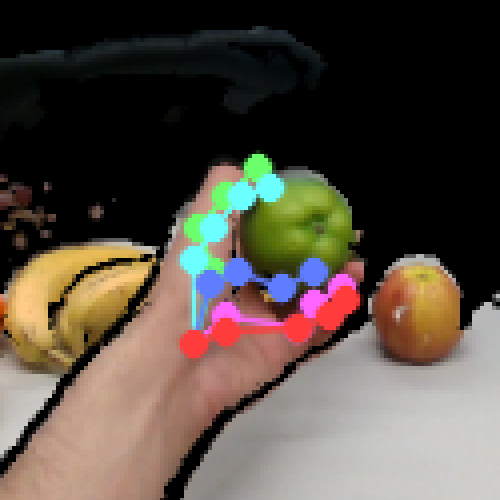}
	\end{subfigure}
	
	\vspace{1.25pt}
	\begin{subfigure}[t]{0.03\textwidth}
		\includegraphics[trim={0 -2.0cm 0 0cm},clip,width=\textwidth]{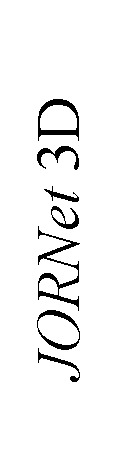}
	\end{subfigure}    
	\hspace{0.05pt}
	\begin{subfigure}[b]{0.155\textwidth}
		\includegraphics[width=\textwidth]{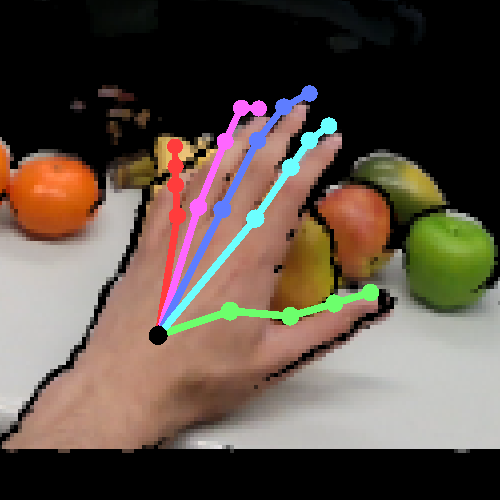}
	\end{subfigure}
	\begin{subfigure}[b]{0.155\textwidth}
		\includegraphics[width=\textwidth]{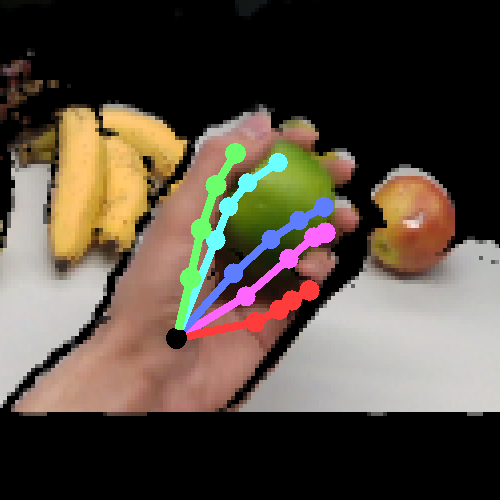}
	\end{subfigure}
	\begin{subfigure}[b]{0.155\textwidth}
		\includegraphics[width=\textwidth]{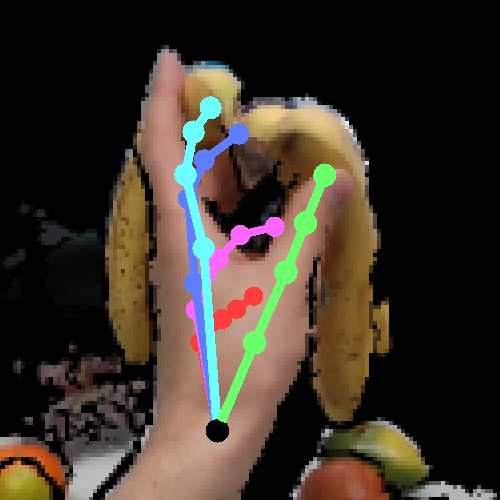}
	\end{subfigure}
	\begin{subfigure}[b]{0.155\textwidth}
		\includegraphics[width=\textwidth]{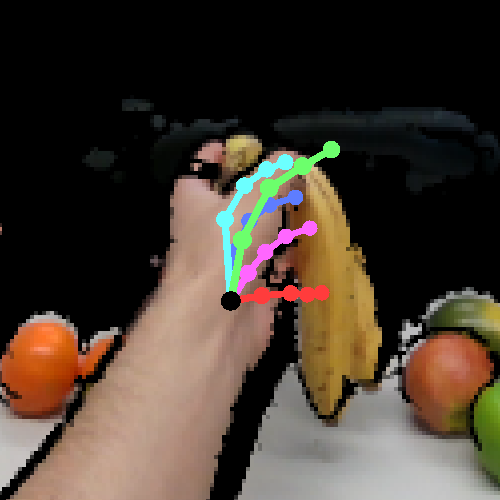}
	\end{subfigure}
	\begin{subfigure}[b]{0.155\textwidth}
		\includegraphics[width=\textwidth]{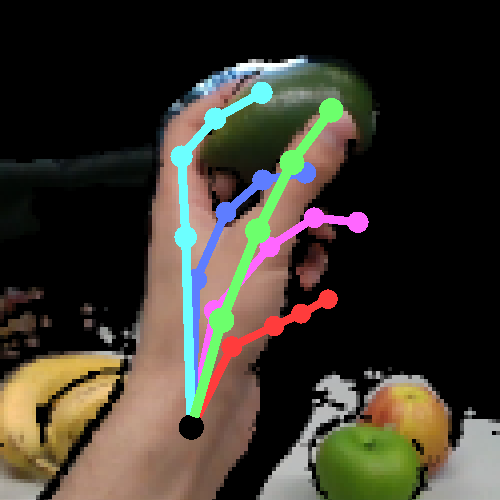}
	\end{subfigure}
	\begin{subfigure}[b]{0.155\textwidth}
		\includegraphics[width=\textwidth]{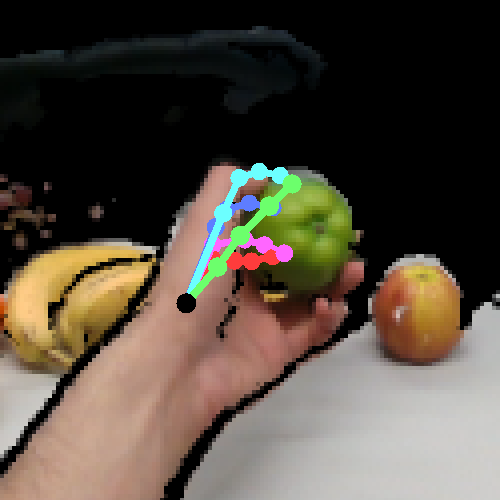}
	\end{subfigure}
	
	\vspace{1.25pt}
	\begin{subfigure}[t]{0.03\textwidth}
		\includegraphics[trim={0 -2.0cm 0 0cm},clip,width=\textwidth]{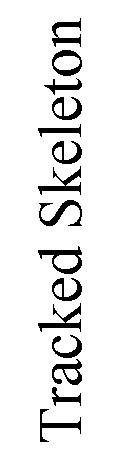}
	\end{subfigure}    
	\hspace{0.05pt}
	\begin{subfigure}[b]{0.155\textwidth}
		\includegraphics[width=\textwidth]{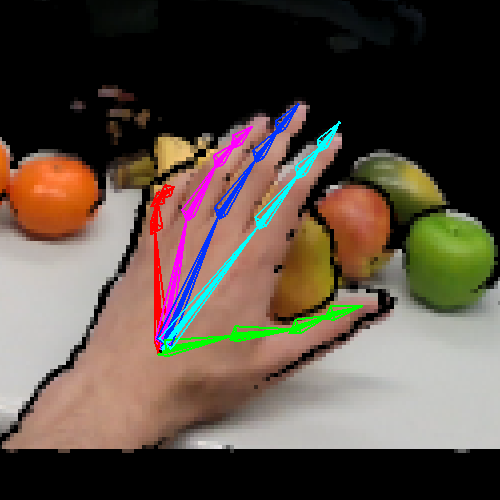}
	\end{subfigure}
	\begin{subfigure}[b]{0.155\textwidth}
		\includegraphics[width=\textwidth]{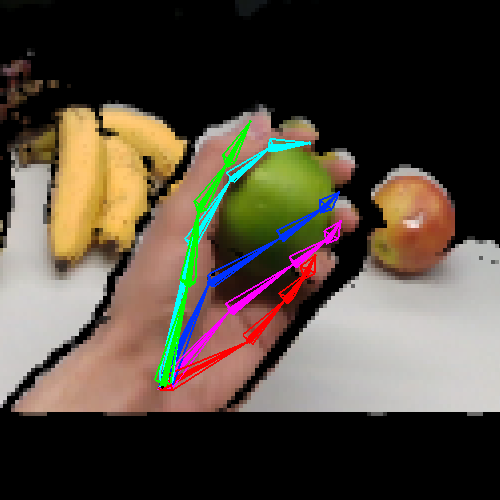}
	\end{subfigure}
	\begin{subfigure}[b]{0.155\textwidth}
		\includegraphics[width=\textwidth]{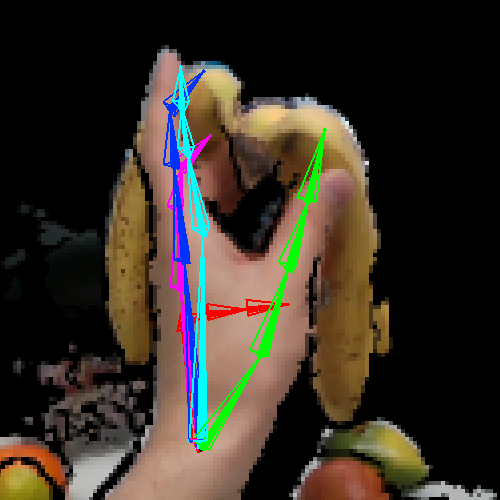}
	\end{subfigure}
	\begin{subfigure}[b]{0.155\textwidth}
		\includegraphics[width=\textwidth]{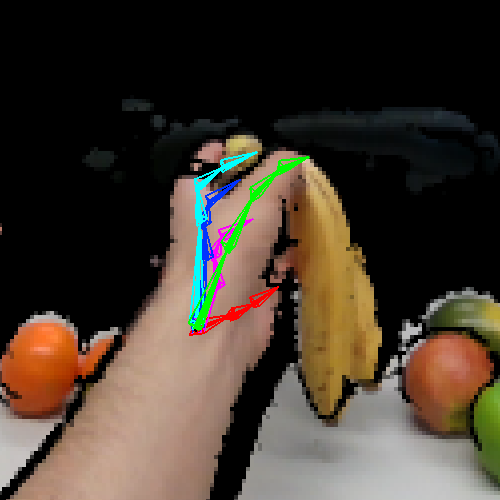}
	\end{subfigure}
	\begin{subfigure}[b]{0.155\textwidth}
		\includegraphics[width=\textwidth]{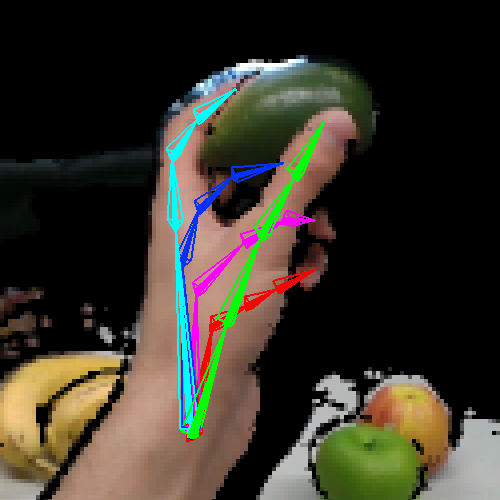}
	\end{subfigure}
	\begin{subfigure}[b]{0.155\textwidth}
		\includegraphics[width=\textwidth]{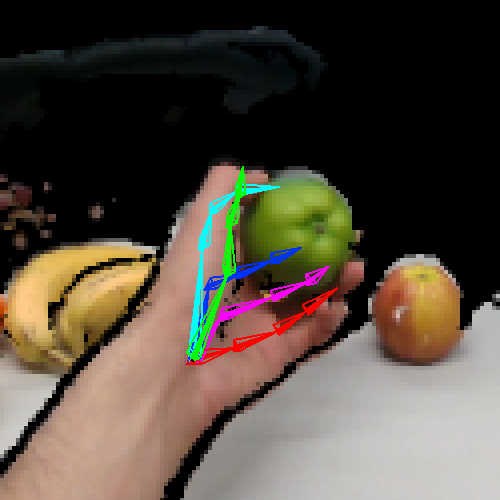}
	\end{subfigure}
	\caption{2D predictions (top row), 3D predictions (middle row) and tracked skeleton (bottom row) on our real annotated sequence \texttt{Fruits}. The combination of 2D and 3D predictions in the tracking framework leads to better results than either of the predictions in isolation.}
	\label{fig:combination_2D_3D}
\end{figure*}

\parahead{Improvement by Combining 2D and 3D Predictions}
Figure~\ref{fig:combination_2D_3D} qualitative depicts the predicted joints by each of the key components of our pipeline --- 2D predictions, 3D predictions, and final tracked skeleton --- on the test sequence \texttt{Fruits}. 
	Note that the modes of failure for the 2D and 3D predictions are different which leads to accurate skeleton tracking even if one kind of prediction is incorrect.
	Thus, the combination of 2D and 3D predictions with the tracking framework consistently produces the best results.

\parahead{Comparisons to the State of the Art}
We were unable to quantitatively evaluate on the only
other existing egocentric hand dataset \cite{rogez_cvpr2015} due to a different sensor currently unsupported by our approach.
Using our method with the \emph{Senz3D} camera requires adaptation of intrinsic camera parameters and noise characteristics for training.
However, to provide a visual comparison to evaluate our method, we recorded sequences that mimic sequences used by Rogez \etal \cite{rogez_cvpr2015}, and show qualitative evaluation in
Figure \ref{fig:qualitative_vs_rogez} and in the supplementary video.
Our method achieves significantly more accurate hand tracking, while running in real time.

\begin{figure*}[h]	
	\centering
	\begin{subfigure}[b]{0.03\textwidth}
		\includegraphics[width=\textwidth]{figures-supplemental/ours_title.jpg}
	\end{subfigure}
	%\hspace{0.05pt}
	\begin{subfigure}[b]{0.155\textwidth}
		\includegraphics[width=\textwidth]{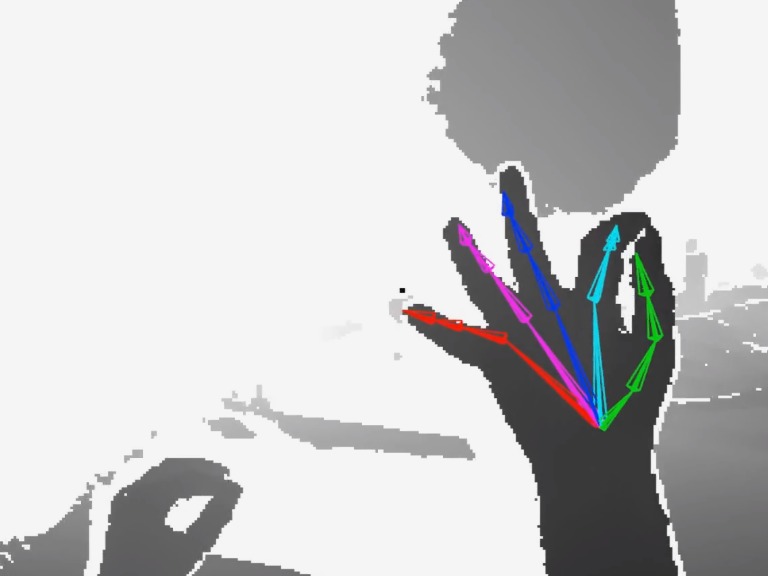}
	\end{subfigure}
	\begin{subfigure}[b]{0.155\textwidth}
		\includegraphics[width=\textwidth]{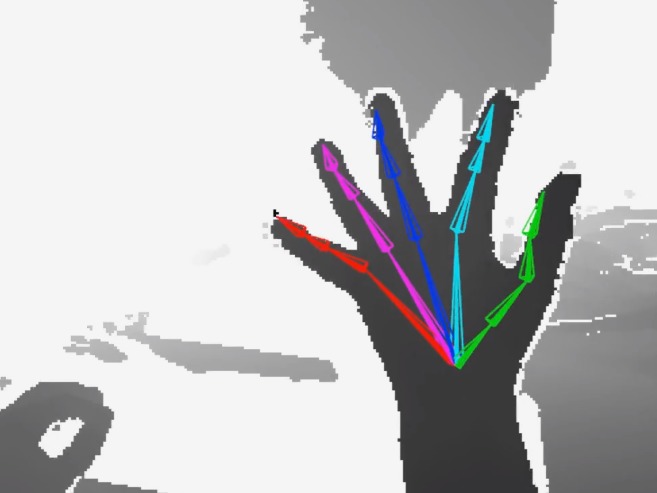}
	\end{subfigure}
	\begin{subfigure}[b]{0.155\textwidth}
		\includegraphics[width=\textwidth]{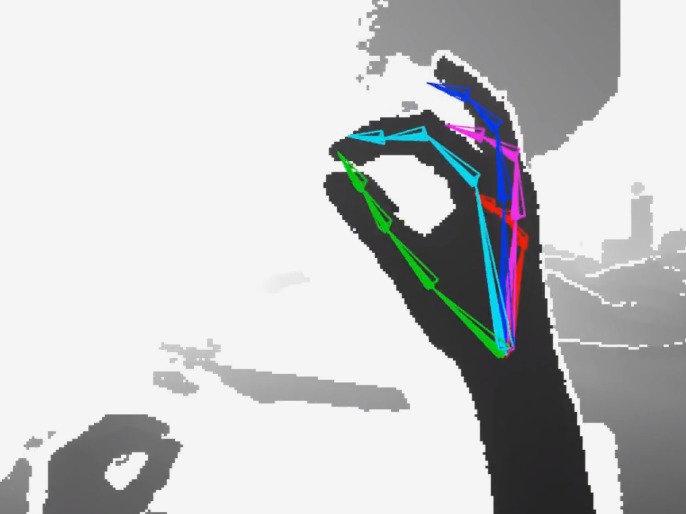}
	\end{subfigure}
	\begin{subfigure}[b]{0.155\textwidth}
		\includegraphics[width=\textwidth]{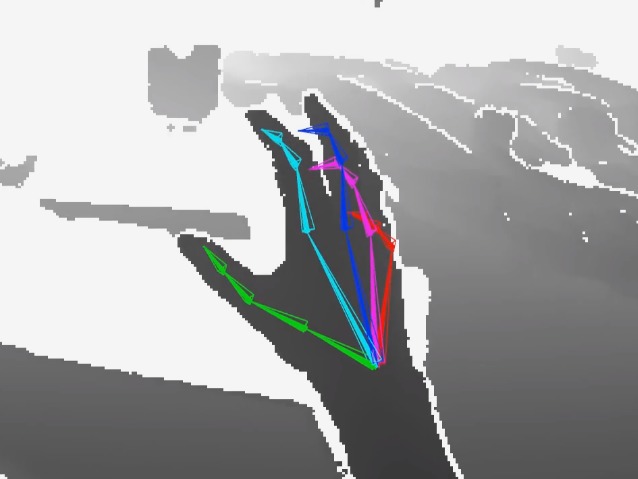}
	\end{subfigure}
	\begin{subfigure}[b]{0.155\textwidth}
		\includegraphics[width=\textwidth]{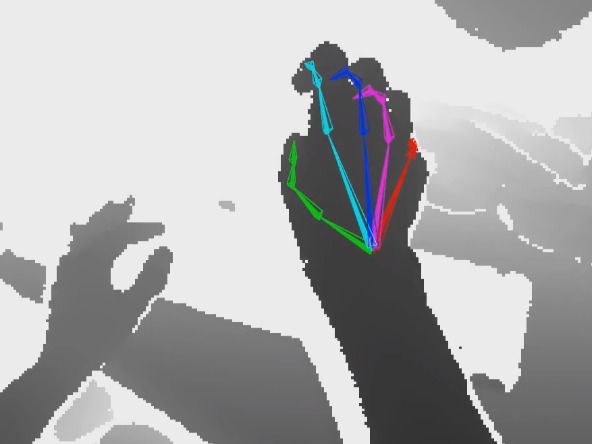}
	\end{subfigure}
	\begin{subfigure}[b]{0.155\textwidth}
		\includegraphics[width=\textwidth]{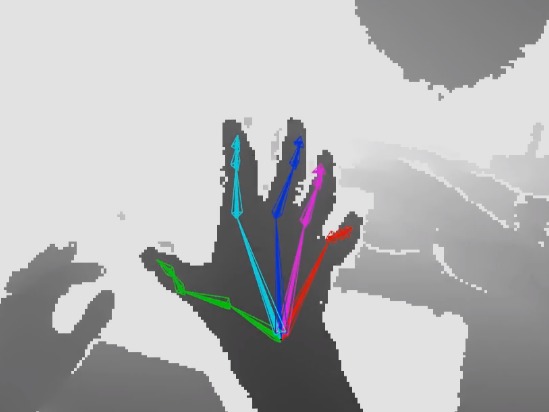}
	\end{subfigure}
	
	\vspace{1.25pt}
	\begin{subfigure}[t]{0.03\textwidth}
		\includegraphics[trim={0 0 0 3.5cm},clip,width=\textwidth]{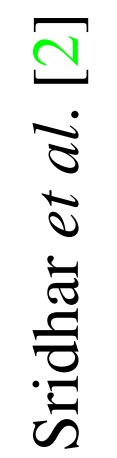}
	\end{subfigure}    
	\begin{subfigure}[t]{0.157\textwidth}
		\includegraphics[width=\textwidth]{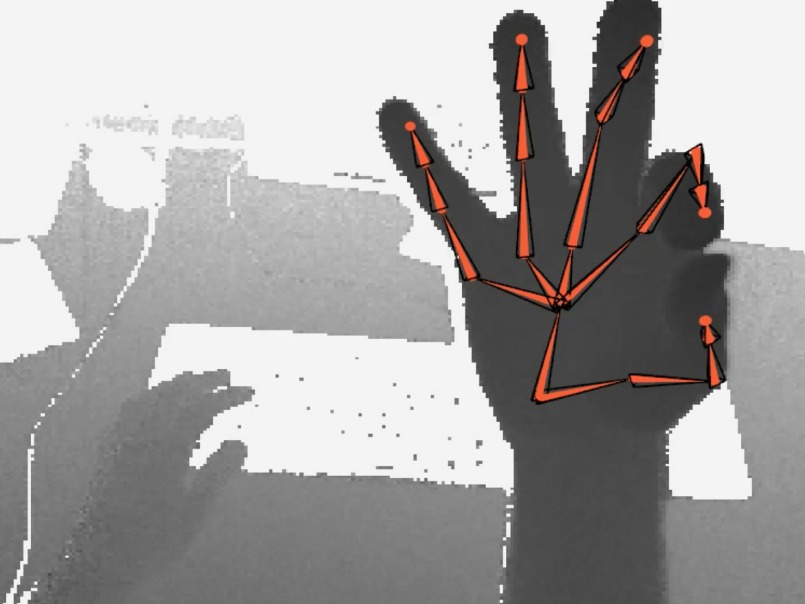}
	\end{subfigure}
	\begin{subfigure}[t]{0.155\textwidth}
		\includegraphics[width=\textwidth]{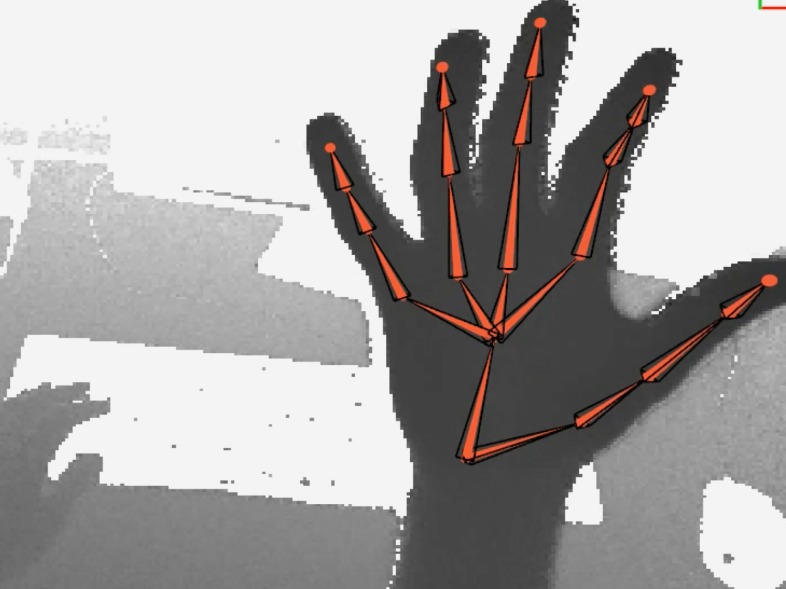}
	\end{subfigure}
	\begin{subfigure}[t]{0.155\textwidth}
		\includegraphics[width=\textwidth]{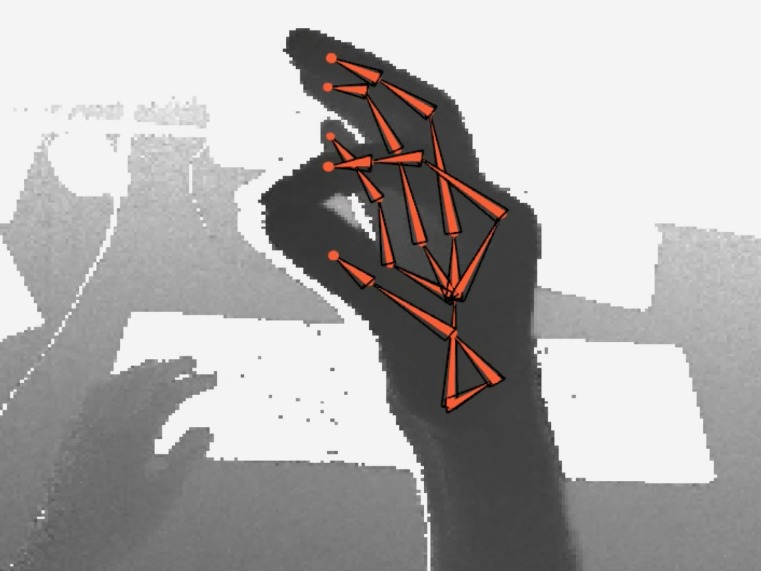}
	\end{subfigure}
	\begin{subfigure}[t]{0.155\textwidth}
		\includegraphics[width=\textwidth]{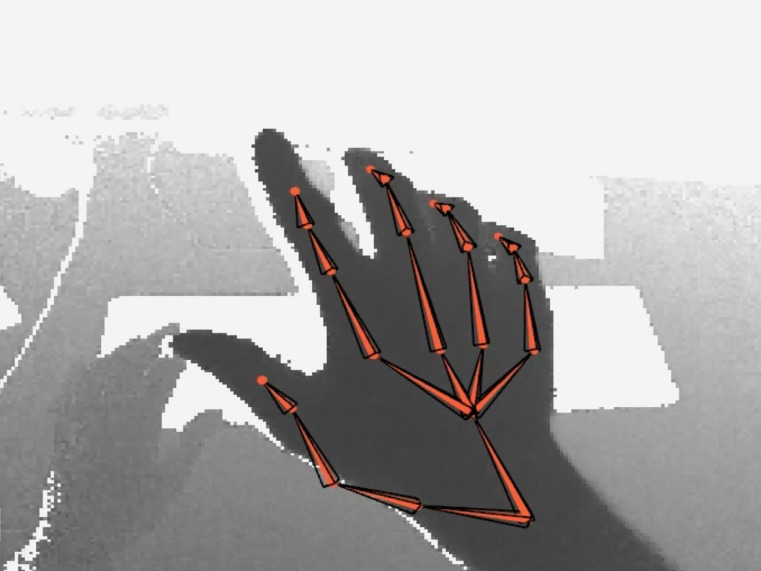}
	\end{subfigure}
	\begin{subfigure}[t]{0.155\textwidth}
		\includegraphics[width=\textwidth]{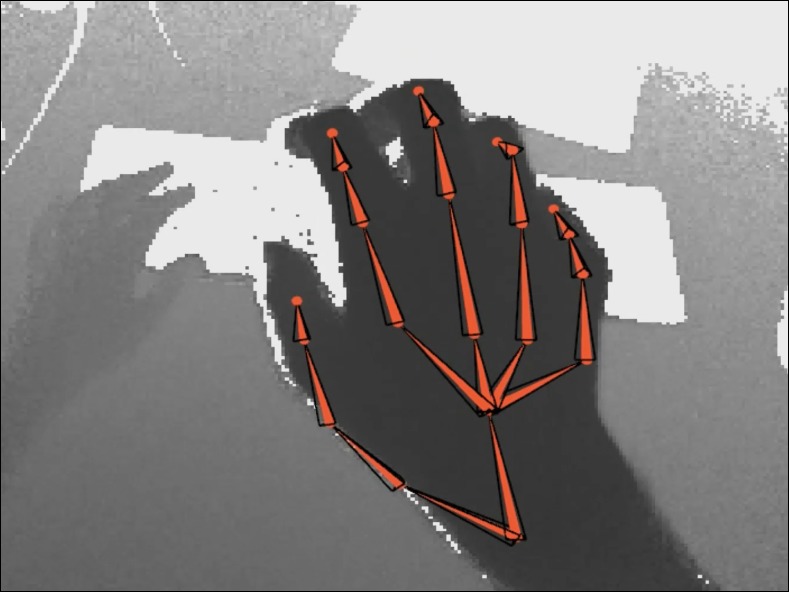}
	\end{subfigure}
	\begin{subfigure}[t]{0.155\textwidth}
		\includegraphics[width=\textwidth]{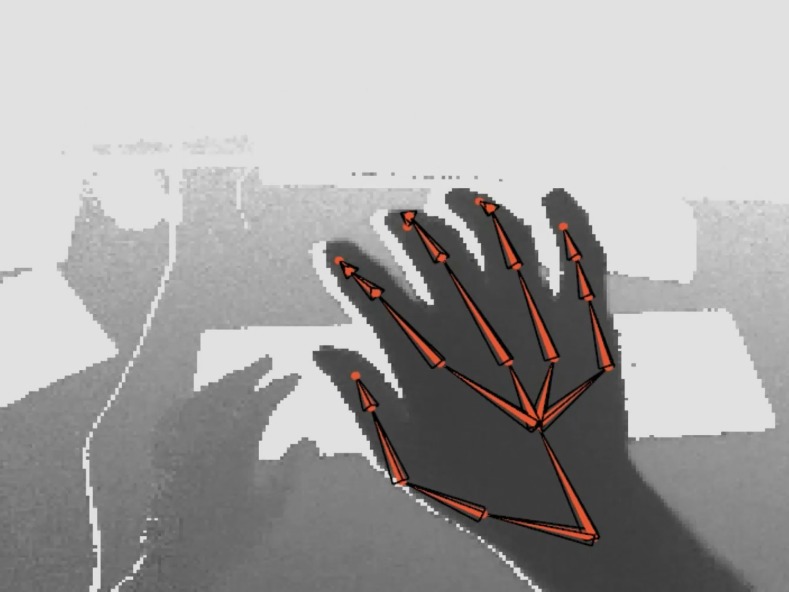}
	\end{subfigure}
	\caption{Qualitative evaluation of our results (top) and Sridhar \etal \cite{sridhar_cvpr2015} (bottom).}
	\label{fig:qualitative_vs_sridhar15}
\end{figure*}

\begin{figure*}[h]	
	\centering
	\includegraphics[width=\textwidth]{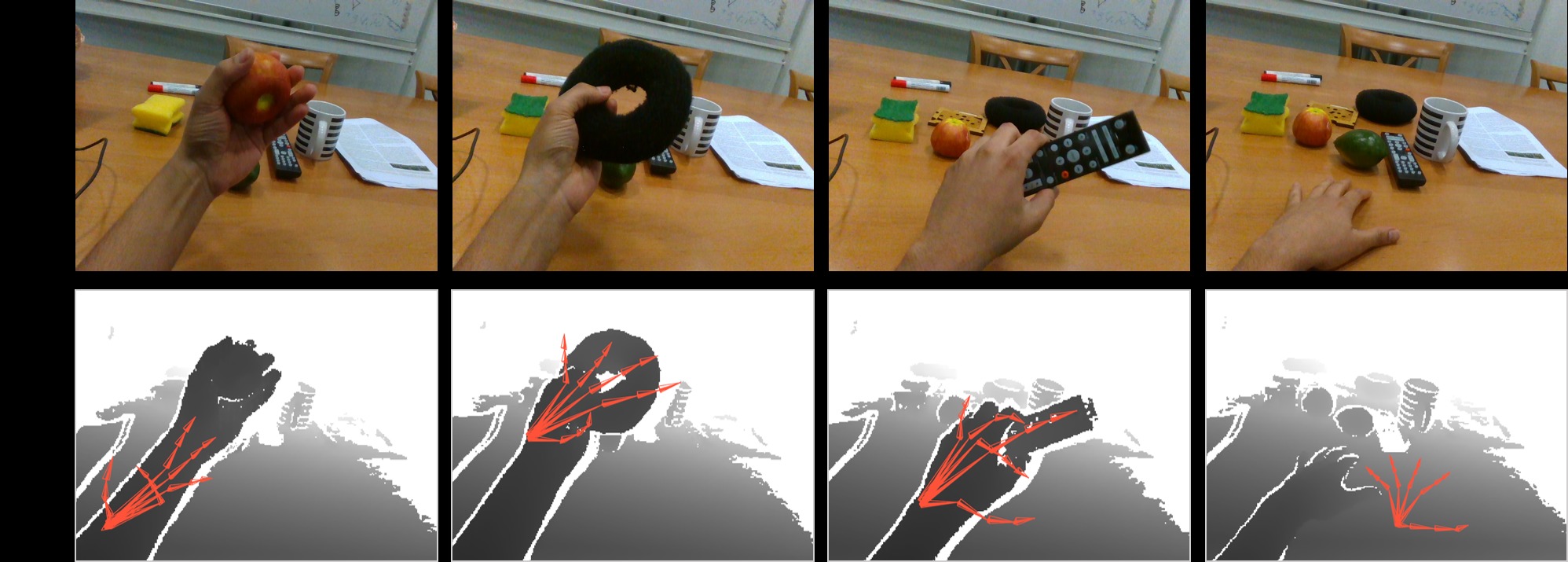}
	\caption{Qualitative results produced by the approach of Sridhar \etal~\cite{sridhar_cvpr2015} on our benchmark dataset \RealBenchmarkName. The close proximity of the arm to the camera and the interaction with objects and the environment leads to catastrophic failures.}
	\label{fig:sridhar15_fail}
\end{figure*}

To show the completely different nature of our problem which cannot be solved by employing state-of-the-art methods for hand tracking in free air, we applied the method of Sridhar \etal~\cite{sridhar_cvpr2015} to our real test sequences from \RealBenchmarkName. Figure~\ref{fig:sridhar15_fail} demonstrates catastrophic failures of the aforementioned approach.
	On the other hand, Figure \ref{fig:qualitative_vs_sridhar15} and the supplemental video show how our method successfully tracks sequences mimicked from the work of Sridhar \etal.
	We achieve comparable results, with improved stability of the hand root position. 

\begin{figure*}[h]	
	\centering
	\includegraphics[width=\textwidth]{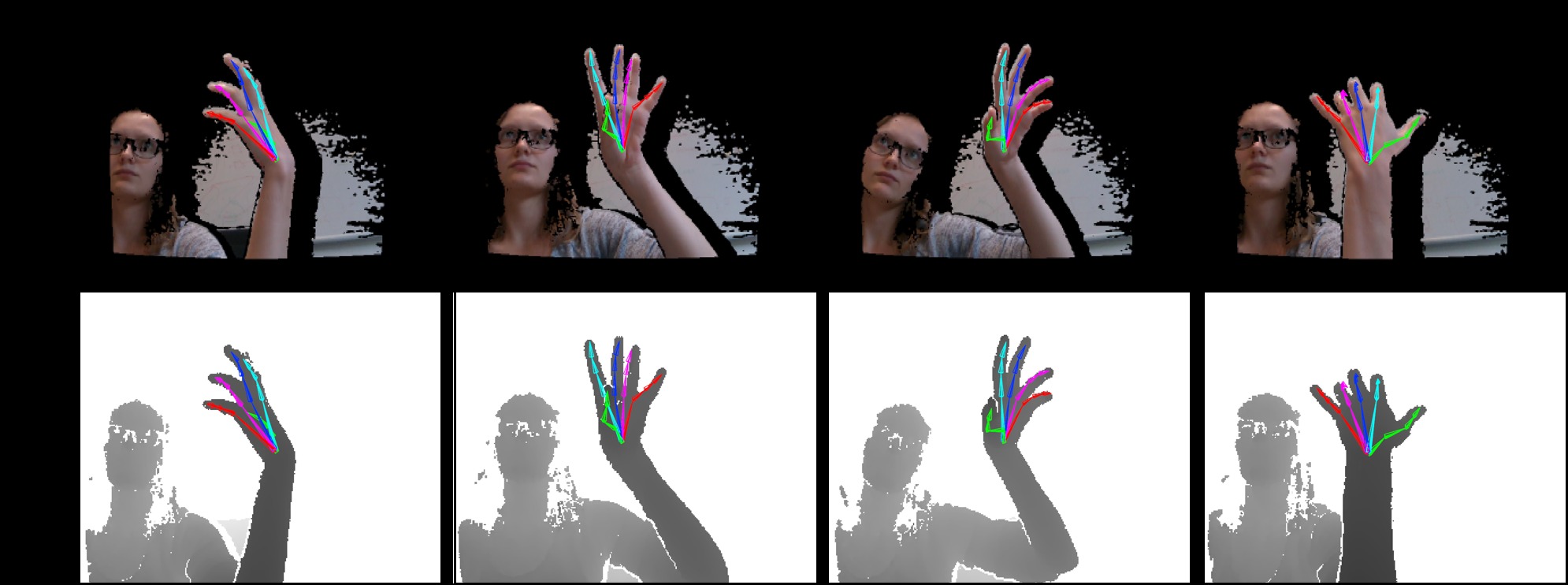}
	\caption{Despite of being trained for egocentric views, our approach in fact generalizes to 3rd-person views. In addition, it is robust to the presence of other skin-colored body parts.}
	\label{fig:results_3rd_person}
\end{figure*}

\parahead{Performance on 3rd-Person Viewpoint}
Even though the machine learning components of our method were only trained on our egocentric dataset \emph{SynthHands}, Figure~\ref{fig:results_3rd_person} demonstrates the generalizability to 3rd-person views. Note that the hand localization step is robust to other skin-colored parts, like faces, in the input.

\begin{figure*}[h]
	\centering							
	\includegraphics[width=\textwidth]{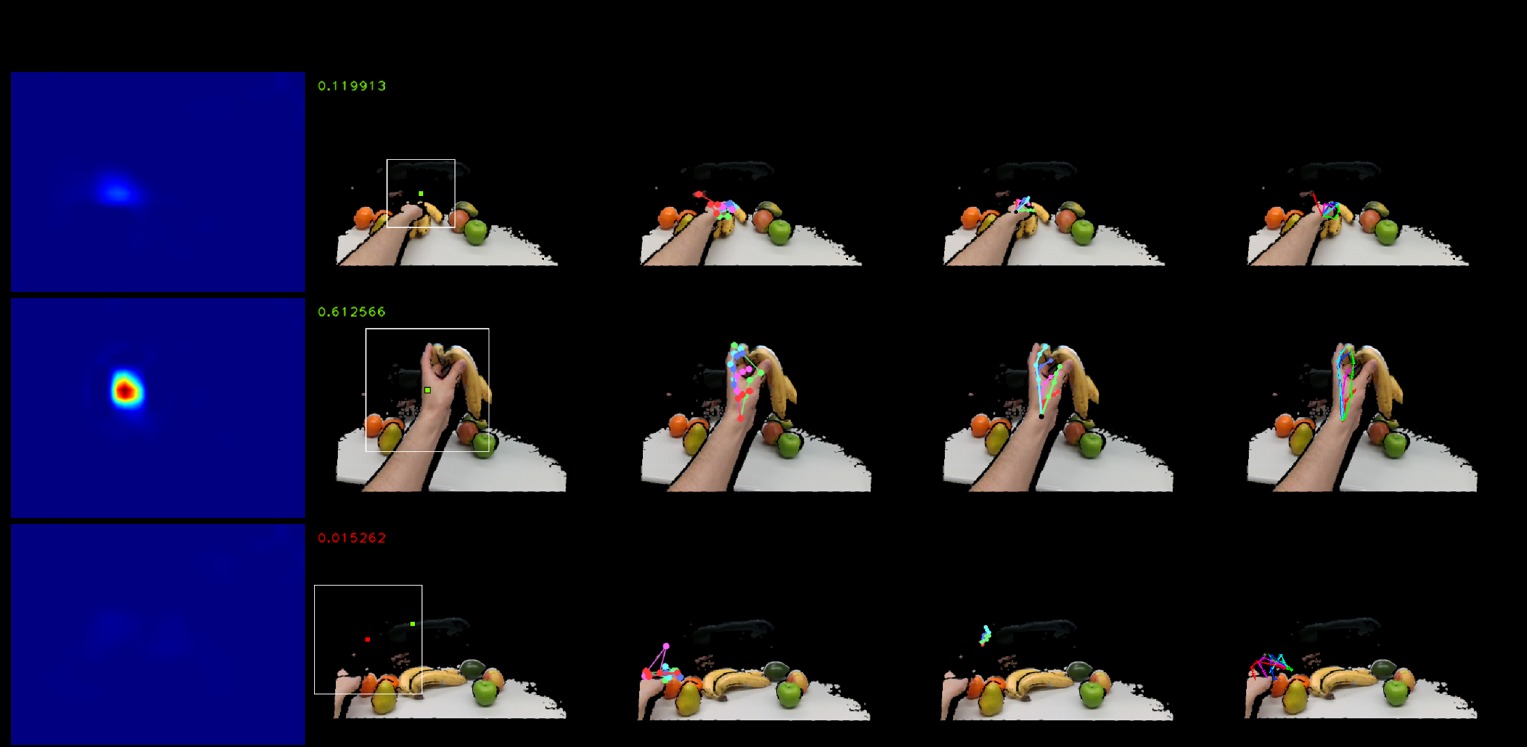}
	\caption{The failure cases of our method can be caused by different intermediate steps. The first two columns show the output from \emph{HALNet}, a heatmap and its maximum location with corresponding confidence. Note that the root stabilization step improved the location in the last row (from green to red) but did not succeed. The third, fourth and fifth column shows 2D predictions, 3D predictions, and the tracked kinematic skeleton, respectively. }
	\label{fig:fail_analysis}
\end{figure*}

\parahead{Analysis of Failure Cases}
Despite the demonstrated success of our approach, we still suffer from limitations that produce erroneous tracking results. 
	Figure~\ref{fig:fail_analysis} depicts the results from several intermediate steps of our method in case of failure. In particular, we show errors due to extreme self occlusions, severe hand and object occlusions, and when the hand is located out of the camera field of view.

{\renewcommand{\arraystretch}{1.2}%
	\begin{table}[b]
		\centering
		\caption{SynthHands Details}
		\label{tab:synthhands_table}
		\begin{tabular}{|l|l|}
			\hline
			\multicolumn{1}{|c|}{\textbf{Mode of Variation}}              & \multicolumn{1}{c|}{\textbf{Amount of Variation}}                                                                \\ \hline \hline
			Pose                                                          & \begin{tabular}[c]{@{}l@{}}63,530 frames of real hand motion, \\ sampled every 5th frame\end{tabular}            \\ \hline
			\begin{tabular}[c]{@{}l@{}}Wrist+Arm \\ Rotation\end{tabular} & \begin{tabular}[c]{@{}l@{}}wrist: sampled from 70 deg. range\\ arm: sampled from a 180 deg. range\end{tabular}   \\ \hline
			Shape                                                         & \begin{tabular}[c]{@{}l@{}}x, y, z scale sampled uniformly in \\  {[}0.8, 1.2{]}; female + male mesh\end{tabular} \\ \hline
			Skin Color                                                    & 2 x 6 hand textures (female/male)                                                                                \\ \hline
			Camera Viewpoints                                             & 5 egocentric viewpoints                                                                                          \\ \hline
			Object Shapes                                                 & 7 objects                                                                                                        \\ \hline
			Object Textures                                               & 145 textures                                                                                                     \\ \hline
			Background Clutter                                            & \begin{tabular}[c]{@{}l@{}}10,000 real images, uniform random \\ u,v offset in {[}-100, 100{]}\end{tabular}      \\ \hline
		\end{tabular}
	\end{table}
}

\section{\emph{SynthHands} Dataset} \label{sec:synthhands}
Table \ref{tab:synthhands_table} shows the modes of variation in the \emph{SynthHands} dataset. Representative frames are shown in Figure \ref{fig:SynthHands}.

\section{CNN Architecture and Training} \label{sec:CNN}
In this section, we explain the network architecture used for \emph{HALNet} and \emph{JORNet} and provide training details. Furthermore, we present experiments which lead to our specific design decisions.
\subsection{Network Design}
The ResNet\cite{he_cvpr2016} architecture has been successfully used for full body pose estimation in previous work \cite{mehta_mlc3d_arxiv16}.
While ResNet50 offers a good tradeoff between speed and accuracy, hand motion is fast and exhibits rapid directional changes. Further, the egocentric camera placement leads to even faster relative motion in the scene.
Therefore, we experimented based on recent investigations into the nature of representations learned by ResNets \cite{greff_highway_iclr16} to get a faster architecture without significantly affecting the accuracy.

\begin{figure*}[h]
	\centering
	\includegraphics[width=0.88\textwidth,page=1]{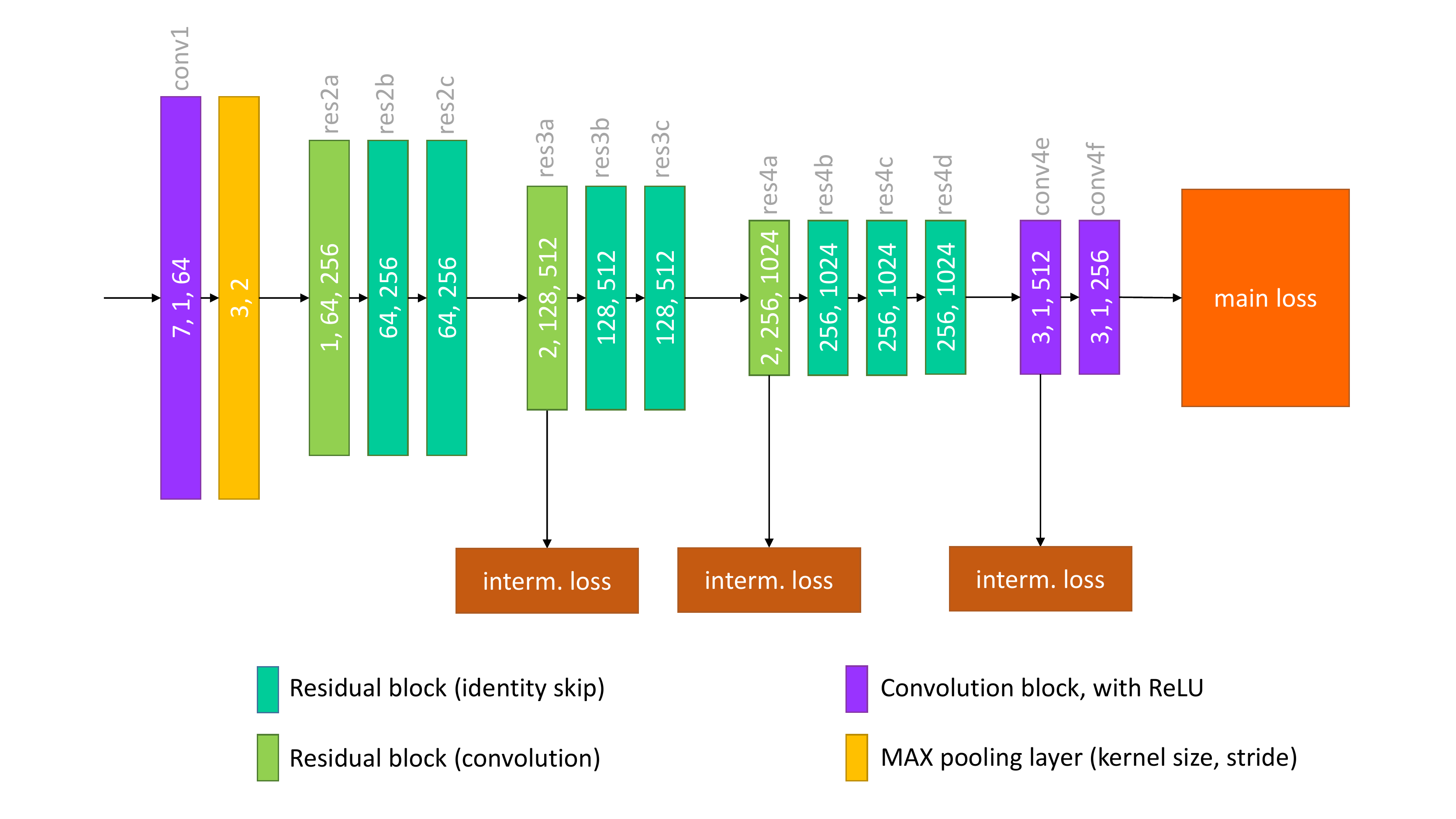}
	\includegraphics[width=0.88\textwidth,page=2,trim={0 9.8cm 0 0},clip]{figures-supplemental/network.pdf}
	\includegraphics[width=0.88\textwidth,page=3]{figures-supplemental/network.pdf}
	\caption{Our network architecture.}
	\label{fig:net_architecture}
\end{figure*}

\parahead{Core Architecture}
Starting from ResNet50, we remove a residual block from level 3, and keep only 4 residual blocks at level 4.
Level 5 is replaced with two $3\times3$ convolution layers with 512 (conv4e) and 256 (conv4f) features and no striding.
Both of these layers also use batch normalization \cite{ioffe2015batch}.
From \emph{conv4f}, A $3\times3$ convolutional stub followed by bilinear upsampling produces the joint location heatmaps, and a fully-connected layer with 200 nodes followed by another fully-connected layer predict the joint location vector.
See Figure \ref{fig:net_architecture} for details.

\begin{figure}[h]	\includegraphics[trim = {3.5cm 8cm 5cm 8cm},clip,width=\columnwidth]{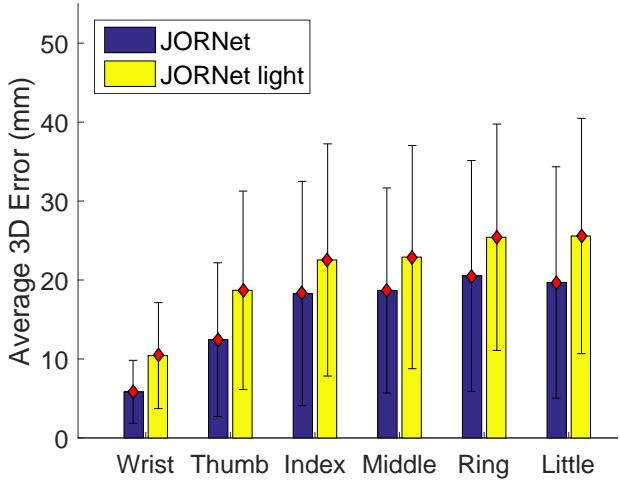}
	\caption{Training \emph{JORNet} to regress heatmaps and local joint positions for all joints instead of only for fingertips and the wrist (\emph{JORNet light}) reduces the error on the \emph{SynthHands} test set.}
	\label{fig:fingertips_wrist}
\end{figure}

The resulting architecture needs 10~ms for a forward pass at resolution 320 x 240 (\emph{HALNet}) and 6~ms at resolution 128 x 128 (\emph{JORNET}) on a Nvidia Pascal Titan X GPU.
This is a significant speed-up compared to the ResNet50 version which needs 18~ms and 11~ms, respectively.
Evaluation on the \emph{SynthHands} test set shows that the drop in accuracy is only marginal.
ResNet50 trained on the hand localization task achieves 2.1~px average error whereas \emph{HALNet} achieves 2.2~px.

\parahead{Intermediate Supervision}
For \emph{HALNet} and \emph{JORNet}, we treat a subset of the feature maps at each of \emph{res3a}, \emph{res4a} and \emph{conv4e} in the networks as the predicted heatmaps, for intermediate supervision \cite{lee_deeply_aistats15}. For \emph{JORNet}, we additionally use the feature maps at the aforementioned stages to predict joint positions for intermediate supervision (see Figure \ref{fig:net_architecture}).

\parahead{Auxiliary Task in \emph{HALNet}}
Predicting heatmaps for all joints as auxiliary task helps \emph{HALNet} to learn the structure of the hand.
This leads to a better performance compared to a version only trained to regress the root heatmap.
On the \emph{SynthHands} test set, regressing heatmaps for all joints instead of only for the root improves the 2D pixel error by \textbf{6.4\%} (from 2.35~px to 2.2~px).

\parahead{Regressing All Joints in \emph{JORNet}}
Although fingertips and wrist location alone provide a strong constraint for the pose of the hand, training \emph{JORNet} to regress the heatmaps and local 3D positions for all joints improves the accuracy.
Figure~\ref{fig:fingertips_wrist} shows the average error for the wrist and all fingertips in 3D on the \emph{SynthHands} test set.
The full \emph{JORNet} version yields a significant increase in performance compared to \emph{JORNet light} which was only trained for wrist and fingertips.

\subsection{Training Details}
We use the Caffe \cite{jia_caffe_ICM} framework for training our networks, using the AdaDelta scheme with momentum set to 0.9 and weight decay to 0.005. Both networks are trained with an input batch size of 16.
For \emph{HALNet}, we use a base learning rate of 0.05 and train for 45k iterations. The input has a spatial resolution of 320x240 px, and the output heatmaps have a resolution of 40x30 px. The main heatmap loss has a loss weight of 1.0, and all intermediate heatmap losses have loss weights of 0.5. 
For \emph{JORNet}, the input has a spatial resolution of 128x128 px. We train with a base learning rate of 0.05, with main heatmap loss weight set at 1.0 and joint position loss weight at 2500. The intermediate heatmap losses have their loss weights set to 0.5 and intermediate joint position loss weights set to 1250. After 45k iterations, the base learning rate is lowered to 0.01, the intermediate heatmap loss weights lowered to 0.1 and the intermediate joint position loss weights lowered to 250, and trained for a further 15k iterations.

\vfill

\begin{figure*}[h!]
	\centering
	\begin{subfigure}[t]{0.155\textwidth}
		\includegraphics[width=\textwidth]{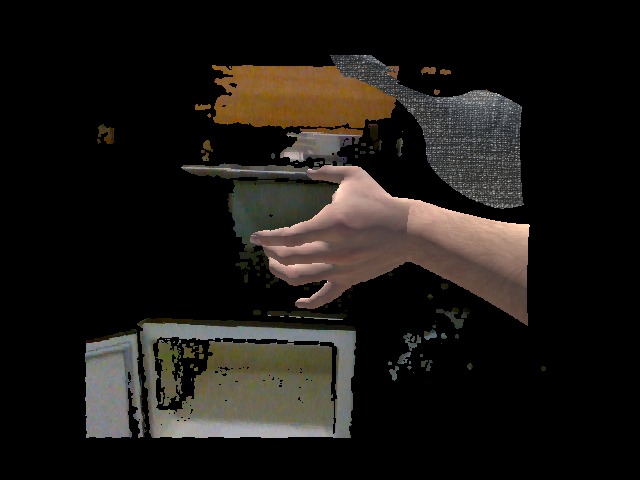}
	\end{subfigure}
	\begin{subfigure}[t]{0.155\textwidth}
		\includegraphics[width=\textwidth]{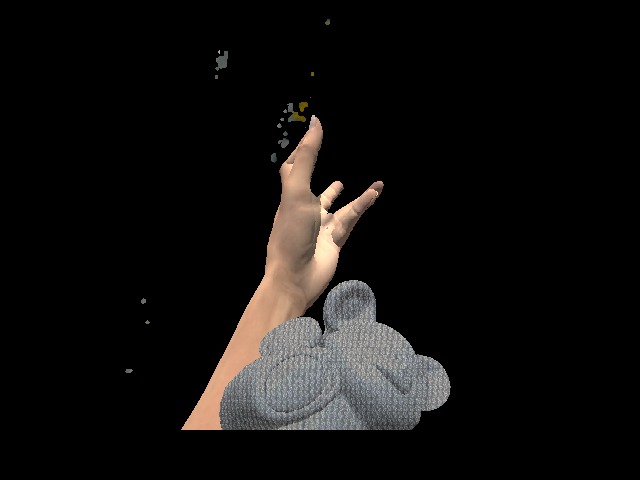}
	\end{subfigure}
	\begin{subfigure}[t]{0.155\textwidth}
		\includegraphics[width=\textwidth]{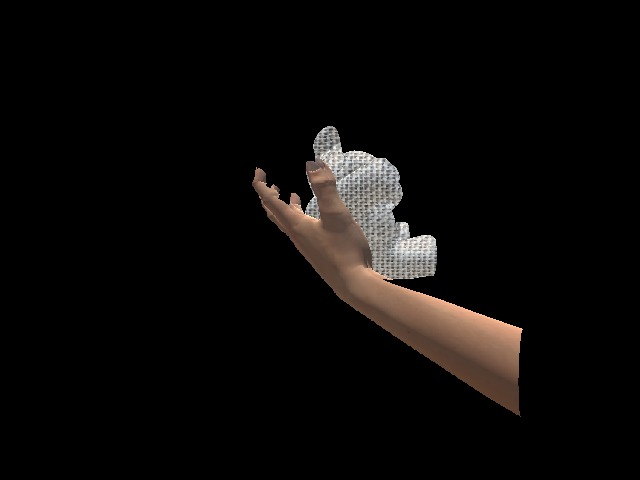}
	\end{subfigure}
	\begin{subfigure}[t]{0.155\textwidth}
		\includegraphics[width=\textwidth]{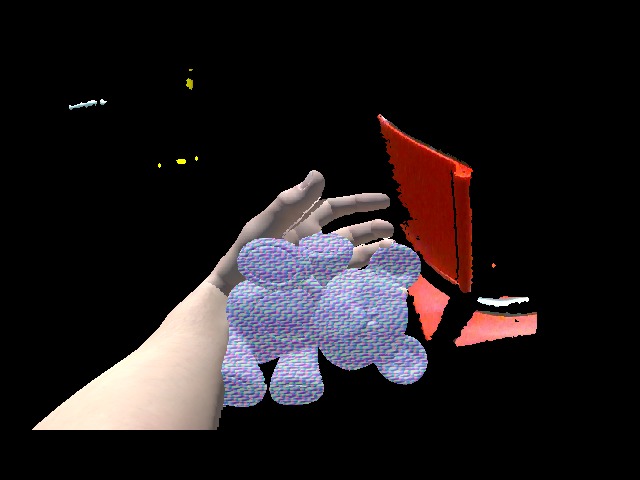}
	\end{subfigure}
	\begin{subfigure}[t]{0.155\textwidth}
		\includegraphics[width=\textwidth]{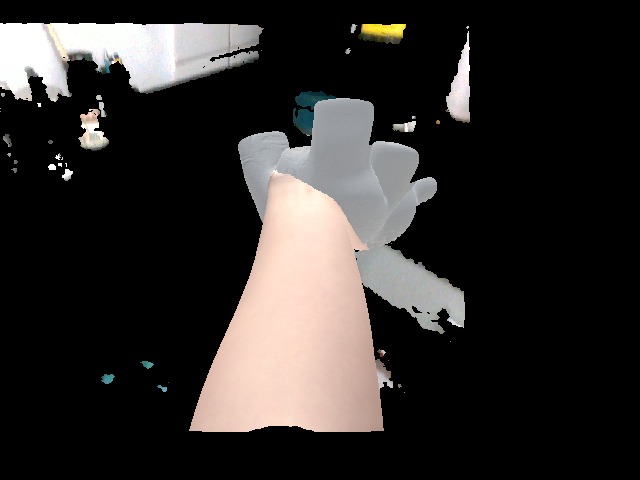}
	\end{subfigure}
	\begin{subfigure}[t]{0.155\textwidth}
		\includegraphics[width=\textwidth]{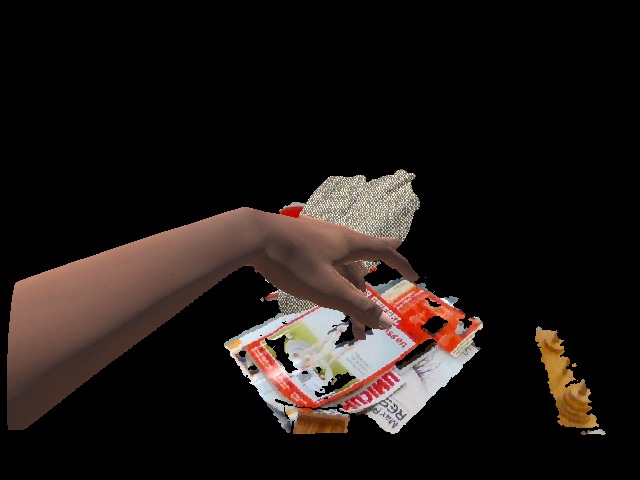}
	\end{subfigure}
	
	\vspace{1.25pt}  
	\begin{subfigure}[t]{0.155\textwidth}
		\includegraphics[width=\textwidth]{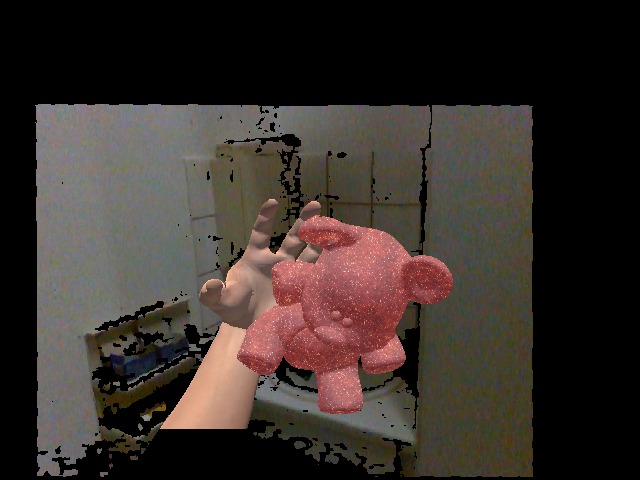}
	\end{subfigure}
	\begin{subfigure}[t]{0.155\textwidth}
		\includegraphics[width=\textwidth]{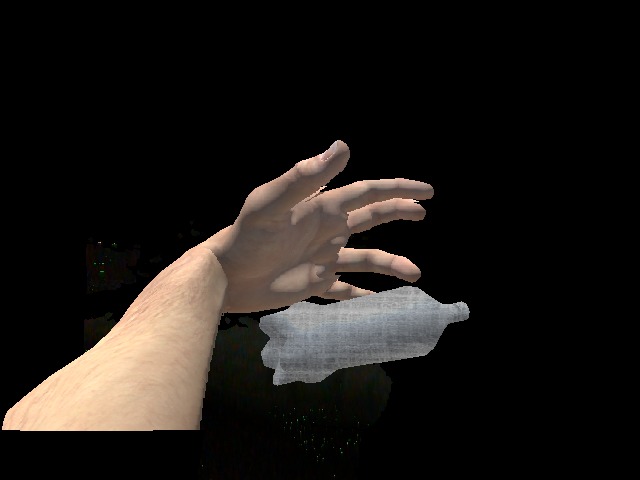}
	\end{subfigure}
	\begin{subfigure}[t]{0.155\textwidth}
		\includegraphics[width=\textwidth]{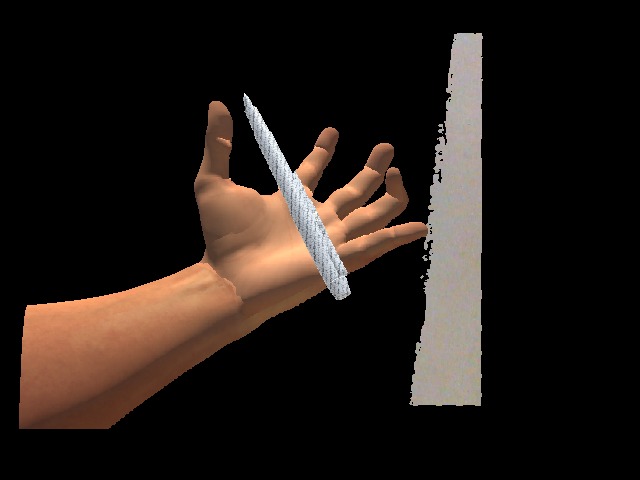}
	\end{subfigure}
	\begin{subfigure}[t]{0.155\textwidth}
		\includegraphics[width=\textwidth]{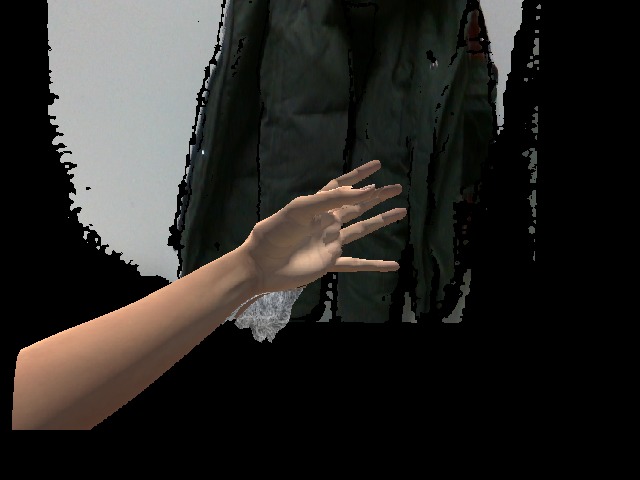}
	\end{subfigure}
	\begin{subfigure}[t]{0.155\textwidth}
		\includegraphics[width=\textwidth]{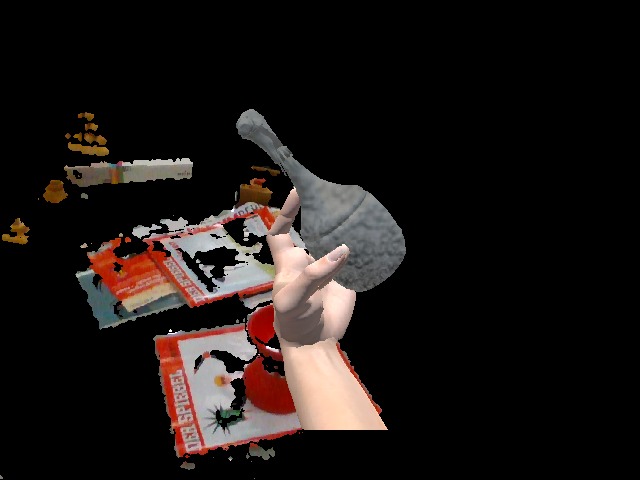}
	\end{subfigure}
	\begin{subfigure}[t]{0.155\textwidth}
		\includegraphics[width=\textwidth]{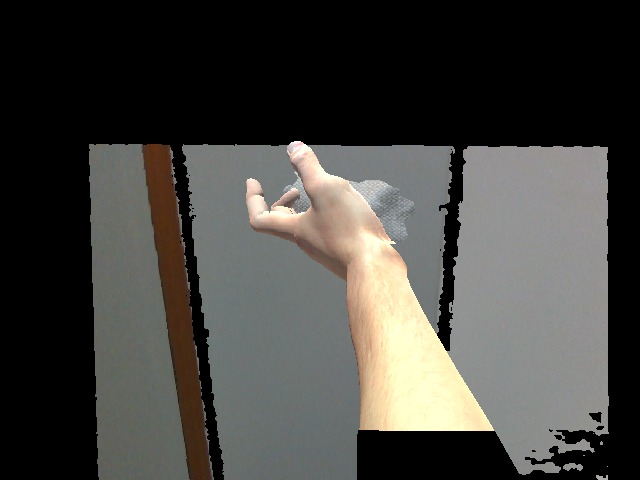}
	\end{subfigure}
	
	\vspace{1.25pt} 
	\begin{subfigure}[t]{0.155\textwidth}
		\includegraphics[width=\textwidth]{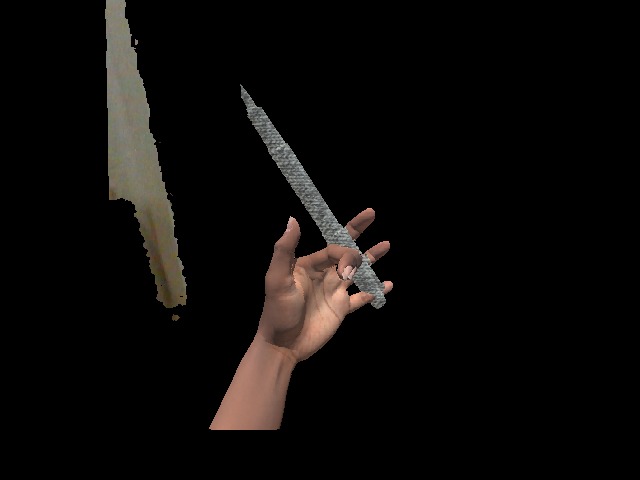}
	\end{subfigure}
	\begin{subfigure}[t]{0.155\textwidth}
		\includegraphics[width=\textwidth]{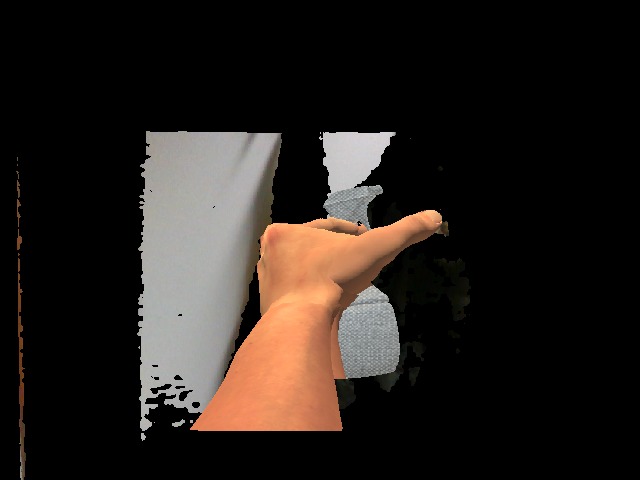}
	\end{subfigure}
	\begin{subfigure}[t]{0.155\textwidth}
		\includegraphics[width=\textwidth]{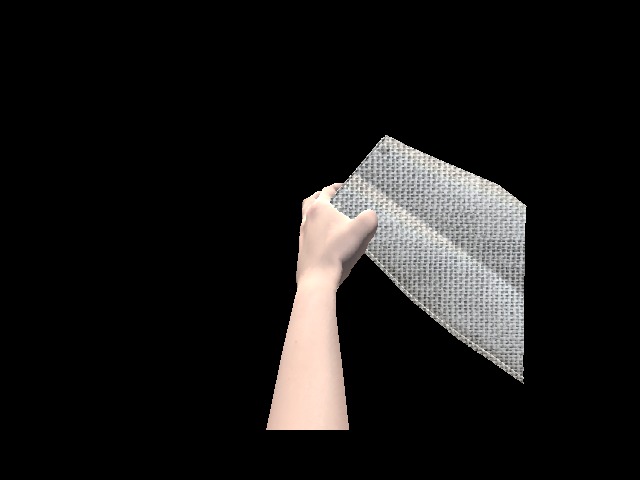}
	\end{subfigure}
	\begin{subfigure}[t]{0.155\textwidth}
		\includegraphics[width=\textwidth]{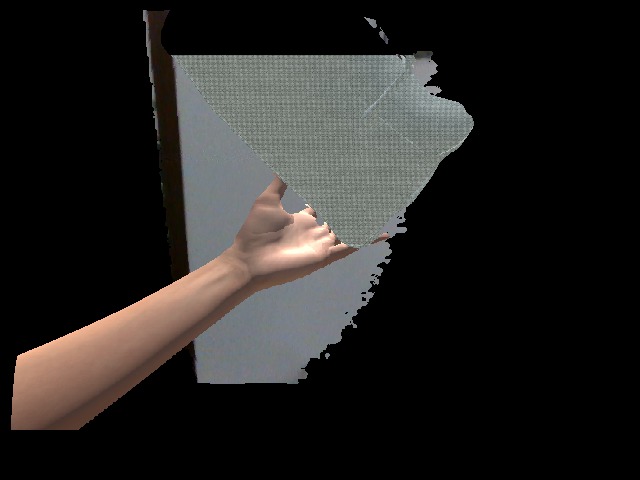}
	\end{subfigure}
	\begin{subfigure}[t]{0.155\textwidth}
		\includegraphics[width=\textwidth]{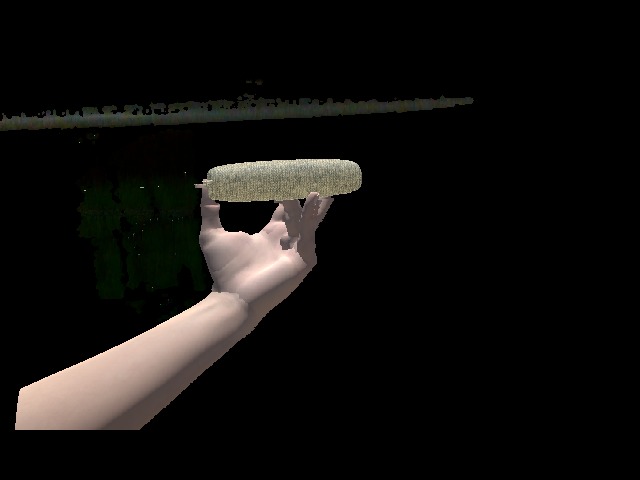}
	\end{subfigure}
	\begin{subfigure}[t]{0.155\textwidth}
		\includegraphics[width=\textwidth]{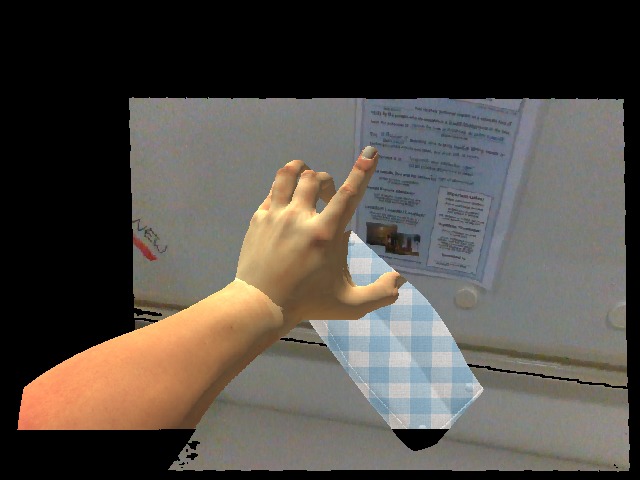}
	\end{subfigure}
	\caption{An example set of random samples taken from our \textit{SynthHands} dataset. See Table \ref{tab:synthhands_table} for description of dataset variability.}
	\label{fig:SynthHands}
\end{figure*}

\begin{figure*}[h]	
	\centering
	\begin{subfigure}[b]{0.03\textwidth}
		\includegraphics[trim={0 0 0 1cm},clip,width=\textwidth]{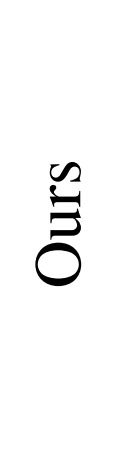}
	\end{subfigure}
	\hspace{0.05pt}
	\begin{subfigure}[b]{0.155\textwidth}
		\includegraphics[width=\textwidth]{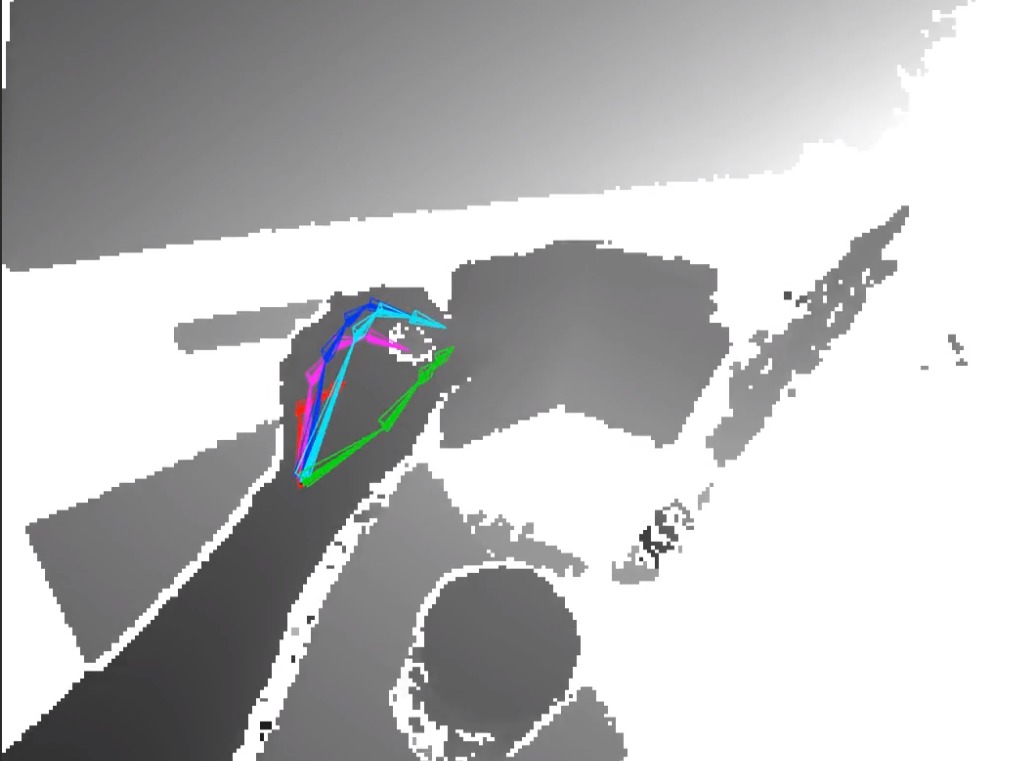}
	\end{subfigure}
	\begin{subfigure}[b]{0.155\textwidth}
		\includegraphics[width=\textwidth]{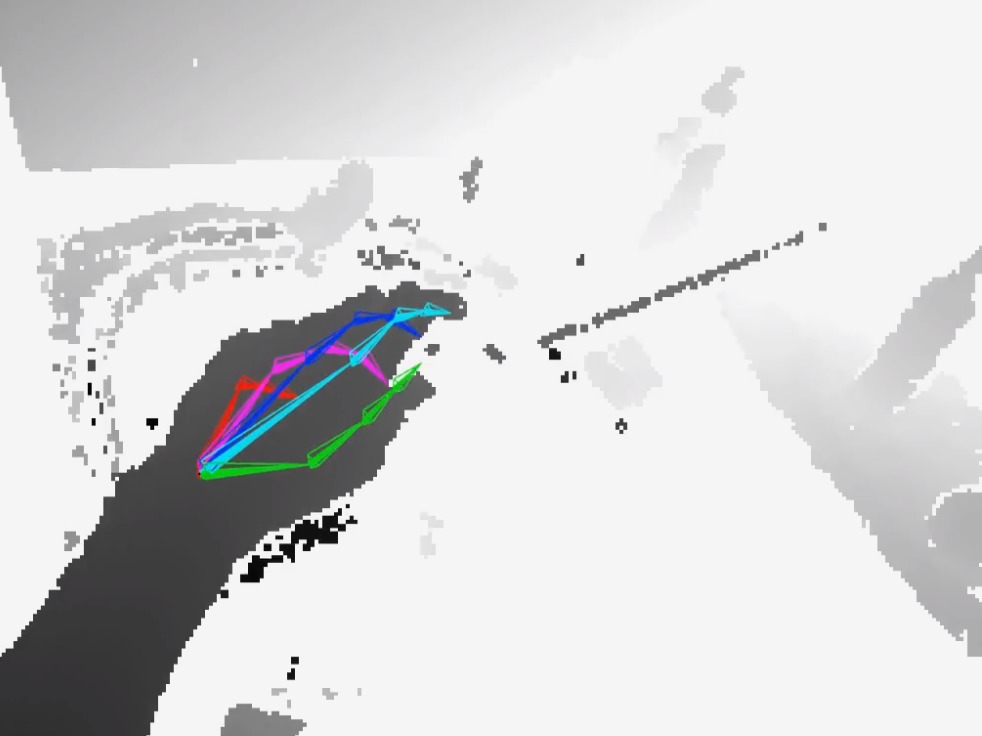}
	\end{subfigure}
	\begin{subfigure}[b]{0.155\textwidth}
		\includegraphics[width=\textwidth]{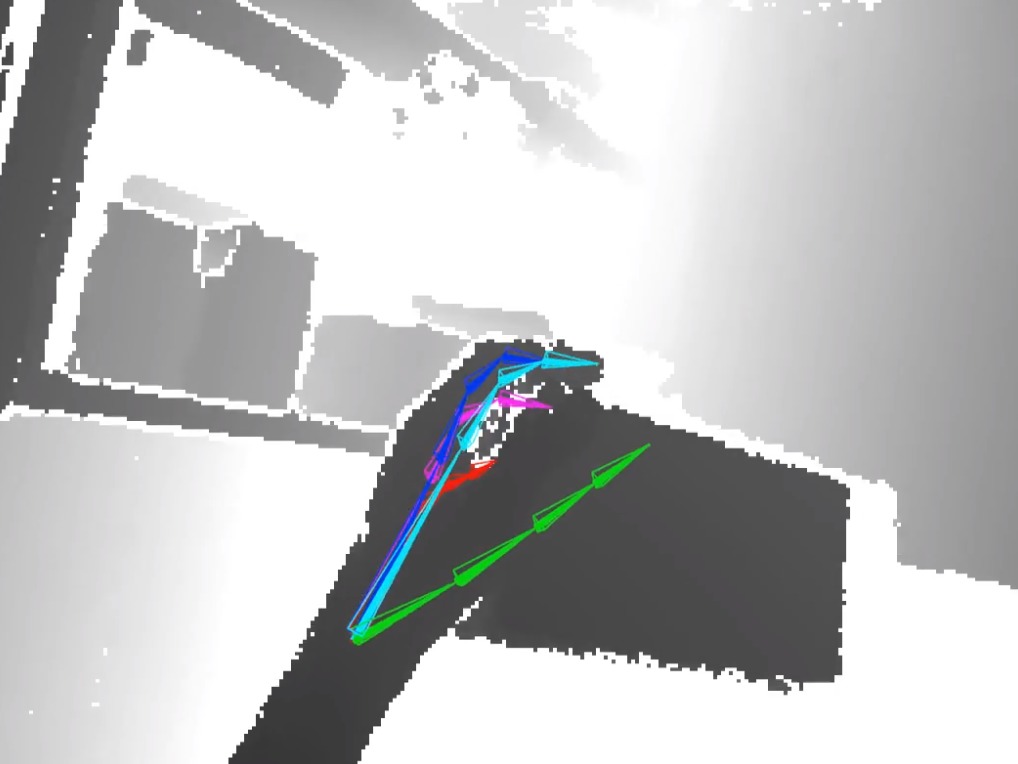}
	\end{subfigure}
	\begin{subfigure}[b]{0.155\textwidth}
		\includegraphics[width=\textwidth]{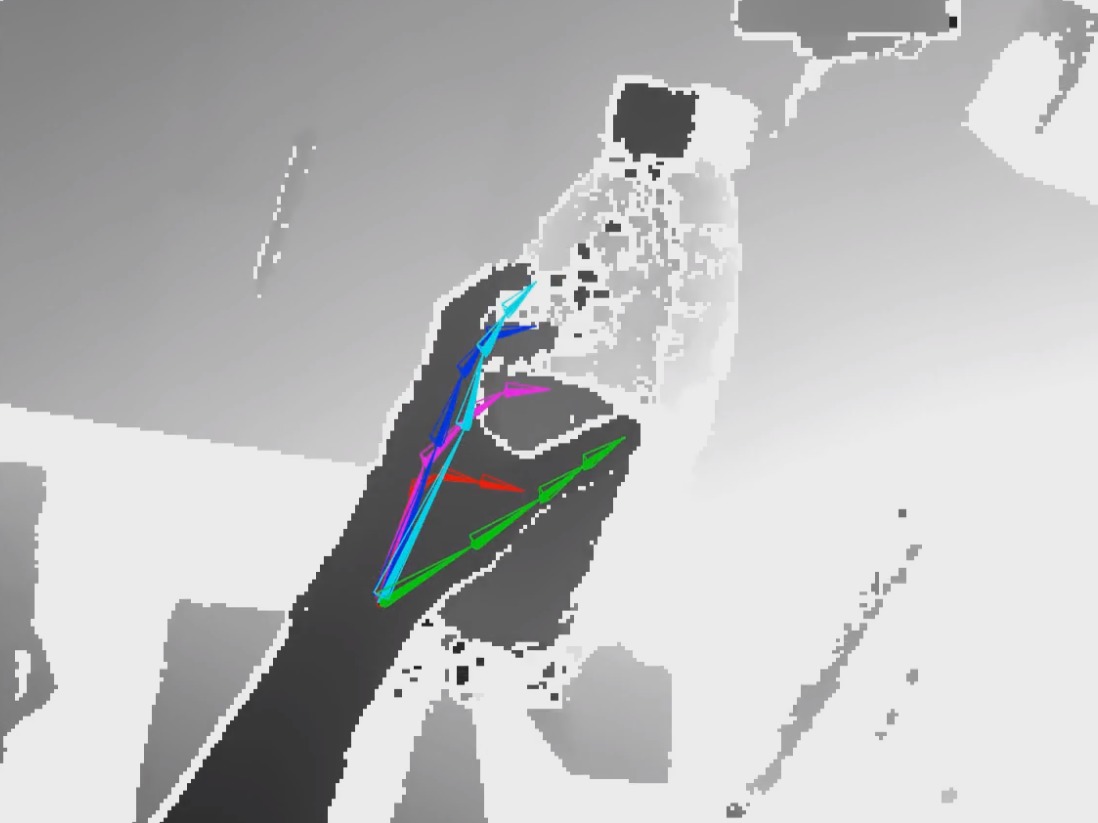}
	\end{subfigure}
	\begin{subfigure}[b]{0.155\textwidth}
		\includegraphics[width=\textwidth]{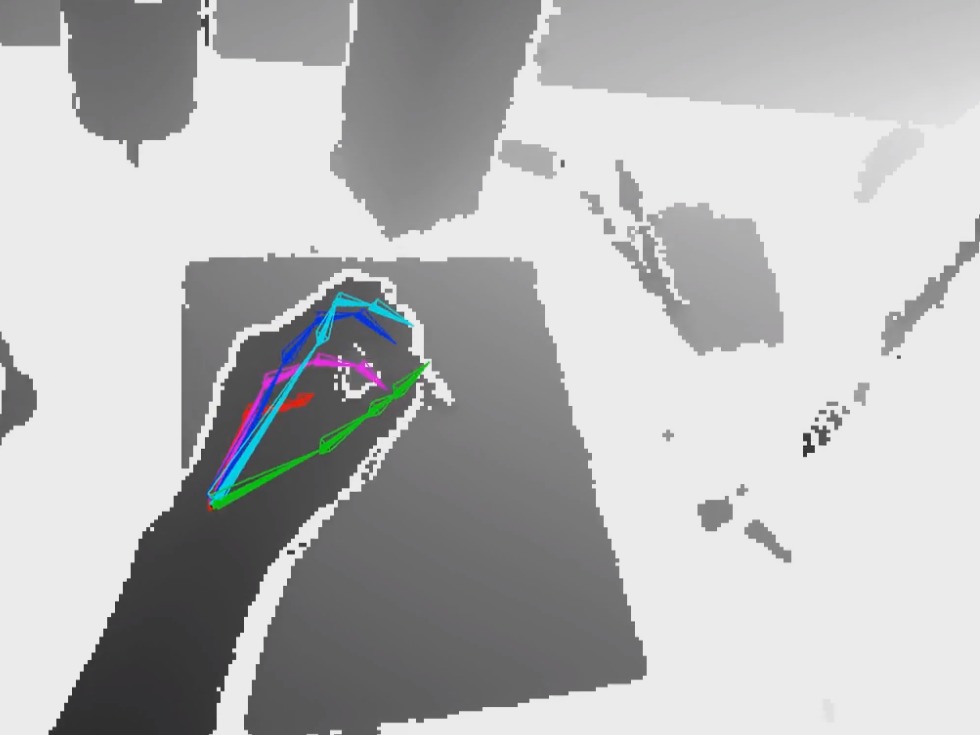}
	\end{subfigure}
	\begin{subfigure}[b]{0.155\textwidth}
		\includegraphics[width=\textwidth]{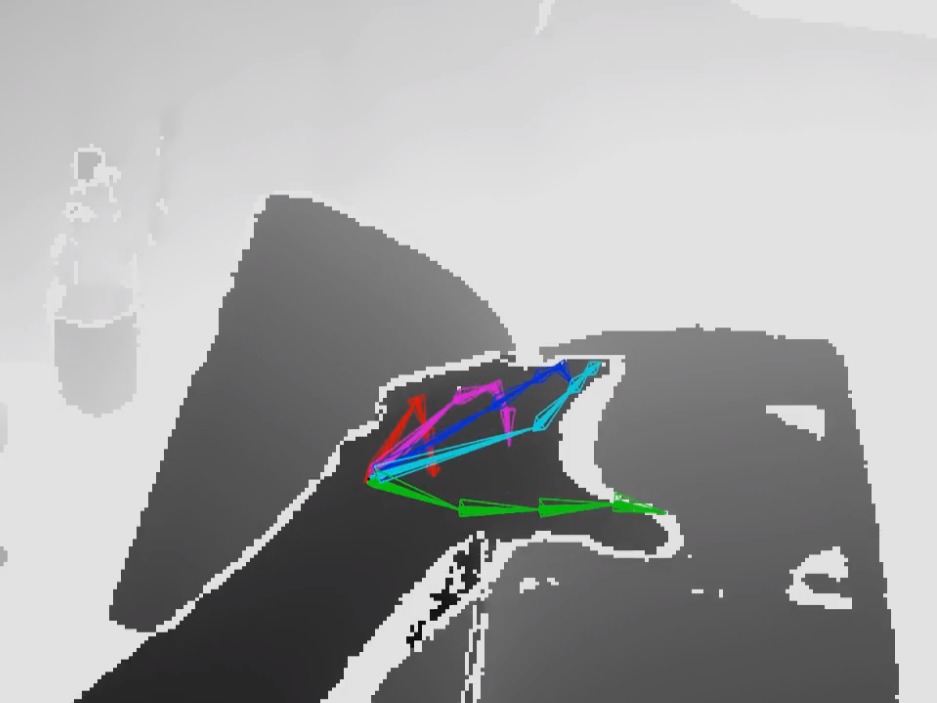}
	\end{subfigure}
	
	\vspace{1.25pt}
	\begin{subfigure}[t]{0.03\textwidth}
		\includegraphics[trim={0 0 0 3.5cm},clip,width=\textwidth]{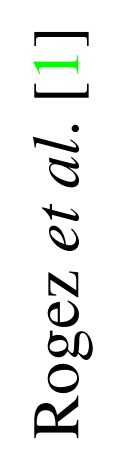}
	\end{subfigure}    
	\begin{subfigure}[t]{0.162\textwidth}
		\includegraphics[width=\textwidth]{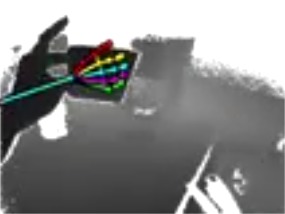}
		\caption{Pinch sponge}
	\end{subfigure}
	\begin{subfigure}[t]{0.155\textwidth}
		\includegraphics[width=\textwidth]{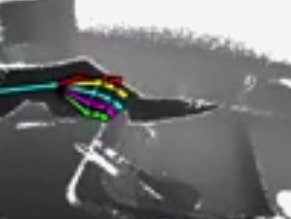}
		\caption{Grab Knife}
	\end{subfigure}
	\begin{subfigure}[t]{0.155\textwidth}
		\includegraphics[width=\textwidth]{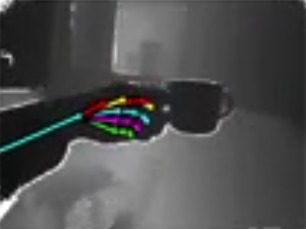}
		\caption{Reach cupboard}
	\end{subfigure}
	\begin{subfigure}[t]{0.155\textwidth}
		\includegraphics[width=\textwidth]{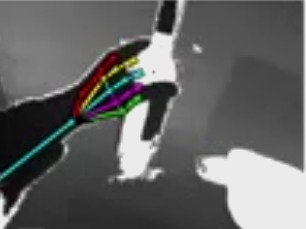}
		\caption{Grab bottle}
	\end{subfigure}
	\begin{subfigure}[t]{0.155\textwidth}
		\includegraphics[width=\textwidth]{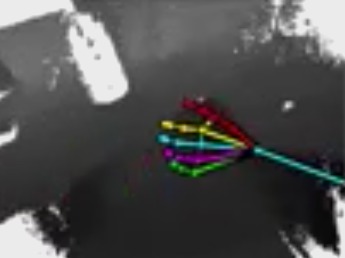}
		\caption{Write in notebook}
	\end{subfigure}
	\begin{subfigure}[t]{0.155\textwidth}
		\includegraphics[width=\textwidth]{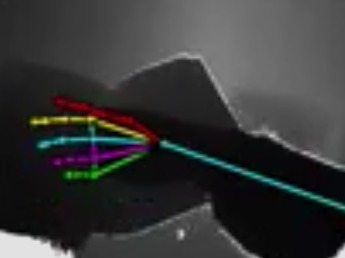}
		\caption{Open book}
	\end{subfigure}
	\caption{Qualitative evaluation of our results (top) and Rogez \etal \cite{rogez_cvpr2015} (bottom). We mimic the motions originally used by \cite{rogez_cvpr2015} because, due to sensor differences (\ie lens intrinsics, etc.), we cannot directly run our trained CNNs on their data.}
	\label{fig:qualitative_vs_rogez}
\end{figure*}

\clearpage
\clearpage

{\small
\bibliographystyle{ieee}
\bibliography{OccludedHands}
}

\end{document}